\documentclass{article}

\usepackage{arxiv}

\usepackage[utf8]{inputenc} 
\usepackage[T1]{fontenc}    
\usepackage{hyperref}       
\usepackage{url}            
\usepackage{booktabs}       
\usepackage{amsfonts}       
\usepackage{nicefrac}       
\usepackage{microtype}      
\usepackage{lipsum}		
\usepackage{graphicx}
\usepackage{natbib}
\usepackage{doi}

\usepackage{amsmath,amssymb,amsfonts}
\usepackage{multirow}
\usepackage{subcaption}
\usepackage{xcolor}

 \title{MARC: Multi-Label Adaptive Retrieval Contrastive Loss for Remote Sensing Images}

\author{ 
Amna Amir \\
  Faculty of Engineering and Natural Sciences \\
  VPA Lab\\
  Sabanci University\\
  Istanbul, Türkiye \\
  \texttt{amna.butt@sabanciuniv.edu} \\
	\And
 Erchan AptoulaARchi \\
   Faculty of Engineering and Natural Sciences \\
  VPA Lab\\
  Sabanci University\\
  Istanbul, Türkiye \\
\texttt{erchan.aptoula@sabanciuniv.edu} \\
}

\date{}

\renewcommand{\headeright}{}
\renewcommand{\undertitle}{}
\renewcommand{\shorttitle}{MARC: Multi-Label Adaptive Retrieval Contrastive Loss for Remote Sensing Images}

\hypersetup{
pdftitle={A template for the arxiv style},
pdfsubject={q-bio.NC, q-bio.QM},
pdfauthor={David S.~Hippocampus, Elias D.~Striatum},
pdfkeywords={First keyword, Second keyword, More},
}

\begin{document}
\maketitle

\begin{abstract}
Semantic overlap among land-cover categories, highly imbalanced label distributions, and complex inter-class co-occurrence patterns constitute significant challenges for multi-label remote-sensing image retrieval. In this article, 
Multi-Label Adaptive Contrastive Learning (MARC) is introduced as an extension of contrastive learning to address them. It integrates label-aware sampling, frequency-sensitive weighting, and dynamic-temperature scaling to achieve balanced representation learning across both common and rare categories.
Extensive experiments on three benchmark datasets (DLRSD, ML-AID, and WHDLD), show that MARC consistently outperforms contrastive-loss based baselines, effectively mitigating semantic imbalance and delivering more reliable retrieval performance in large-scale remote-sensing archives. Code, pretrained models, and evaluation scripts will be released at \url{https://github.com/Amna-128/MARC} upon acceptance. 
\end{abstract}


\section{Introduction}
The evolution of remote sensing technologies has significantly increased the daily volume of satellite imagery acquisition, resulting in vast archives that require efficient data management solutions\cite{di2023bigdata}.  
Content-based image retrieval (CBIR) addresses this challenge by retrieving images from large databases based on their visual content rather than on textual or metadata descriptions\cite{datta2008image}. In this process, discriminative features are extracted from both the query images and the images stored in the database, after which similarity measures are computed to retrieve the most relevant results.

However, the application of CBIR to remote sensing imagery is particularly challenging \cite{zhou2023remote}, since remote sensing scenes are characterized by intricate spatial structures, highly variable object scales, and diverse spectral characteristics, where each image is often composed of multiple land-cover types or semantic classes that are spatially co-occurring and overlapping. Consequently, \textit{multi-label} CBIR, where each sample is associated with an arbitrary number of labels, is considered a more realistic paradigm \cite{sumbul2022relevant,sumbul2022informative,imbriaco2022toward}.

Notable approaches in this regard include fully convolutional networks extracting pixel-level features \cite{shao2020multilabel}, deep hashing methods compressing high-dimensional features into compact binary codes for efficient retrieval \cite{Han2024Hash, Li2020} and graph-based strategies capturing inter-class relationships \cite{Kang2021GraphRelationNetwork}. Nevertheless, all aforementioned methods were limited by predefined similarity measures and the static nature of graph structures. Which is why, \textit{contrastive learning} (CL) \cite{chen2020simclr}, as a data-driven representation learning paradigm has become immensely a popular for retrieval purposes \cite{liu2025dynamic}. Within CL, feature embeddings are learned by maximizing the agreement between semantically similar samples while minimizing similarity between dissimilar ones, leading to compact intra-class and well-separated inter-class representations. 

There are various reported cases of CL based CBIR solutions for remote sensing \cite{huang2023sclfusion}, showing its ability to capture semantic similarity relations. However, most existing approaches have been developed under single-label assumptions, where each image is assumed to represent a single concept \cite{khosla2021supcon}. Attempts to extend CL to multi-label  settings have been made \cite{zhang2024mulsupcon, huang2024similarity, audibert2024survey}, where the main challenge arises from the ambiguities in defining positive and negative pairs given the multitude of labels per sample. Still, they ignore category co-occurrence frequencies and class balances, thus limiting their potential.

In this article, a multi-label extension to supervised CL is proposed, named \textit{Multi-Label Adaptive Contrastive Learning} (MARC), aiming for balanced and relationship-aware representation learning. It incorporates adaptive mechanisms to adjust the learning process according to label relationships and category frequency. MARC supports label-aware sampling for balanced pair formation, frequency-sensitive weighting to highlight rare categories, and dynamic-temperature scaling to adjust similarity sensitivity based on semantic overlap.
Extensive experiments on three benchmark datasets (DLRSD, ML-AID, and WHDLD), show that MARC consistently outperforms CL based baselines, effectively mitigating semantic imbalance and delivering more reliable retrieval performance in large-scale remote-sensing archives.

\section{Related Work}
\subsection{Single-Label Remote Sensing Image Retrieval (SLRSIR)}
SLRSIR has been developed to retrieve remote sensing scenes assumed to represent a single semantic concept. 
Among the large volume of existing methods in this regard, the most notable include deep metric and triplet-based learning networks, that have been used to optimize intra-class compactness and inter-class separability \cite{Zhang2021, Cao2019, Zhao2022}. In particular, dual-anchor triplet loss and global-aware ranking strategies have been reported to strengthen the discriminative capability of learned embeddings \cite{Fan2021}. A learnable joint spatial–spectral transformation has also been proposed to enhance image representation \cite{Wang2021a}.

Moreover, attention and graph-based frameworks have also been studies intensively. Specifically, 

multi-attention fusion networks with dilated convolution and label smoothing have been developed to capture multi-scale contextual cues while suppressing redundant information \cite{Wang2022}. In addition, attention-driven graph convolutional networks have been proposed to model spatial dependencies among image patches, leading to context-aware representations that significantly improve retrieval robustness \cite{Chaudhuri2022}. A self-attention feature metric learning framework has also been introduced to jointly leverage attention mechanisms and metric learning, effectively enhancing the discriminative power of learned features and semantic alignment \cite{Wu2023}.

As far as efficiency is concerned, large-scale retrieval systems have often adopted hashing and meta-learning paradigms. In detail, asymmetric hash code learning frameworks have been employed to generate discriminative binary codes without compromising accuracy \cite{Song2022}. Meta hashing strategies have also been developed to adapt hash functions dynamically to new data distributions \cite{Tang2022}. Adaptive hash-code balancing frameworks have addressed in particular the imbalance and difficulty spectrum among image pairs in SLRSIR, resulting in more robust retrieval over varied categories \cite{Wang2023}. These developments have provided a balance between retrieval speed and semantic precision, making them suitable for real-world SLRSIR systems. 

Furthermore, plasticity–stability preserving multi-task frameworks have been proposed to retain previously acquired representations while adapting to new data \cite{Sumbul2022}. Besides, centripetal-intensive deep hashing \cite{Li2025} and deep attention hashing with distance-adaptive ranking have advanced hybrid learning between metric and hashing paradigms, improving semantic compactness and retrieval precision \cite{Zhang2023}. 
However, despite the strong progress of SLRSIR, their assumption of one dominant category per image has restricted their applicability to complex remote sensing scenes that typically contain multiple semantic concepts.

\begin{figure*}[!ht]
\begin{center}
    \includegraphics[width=\linewidth]{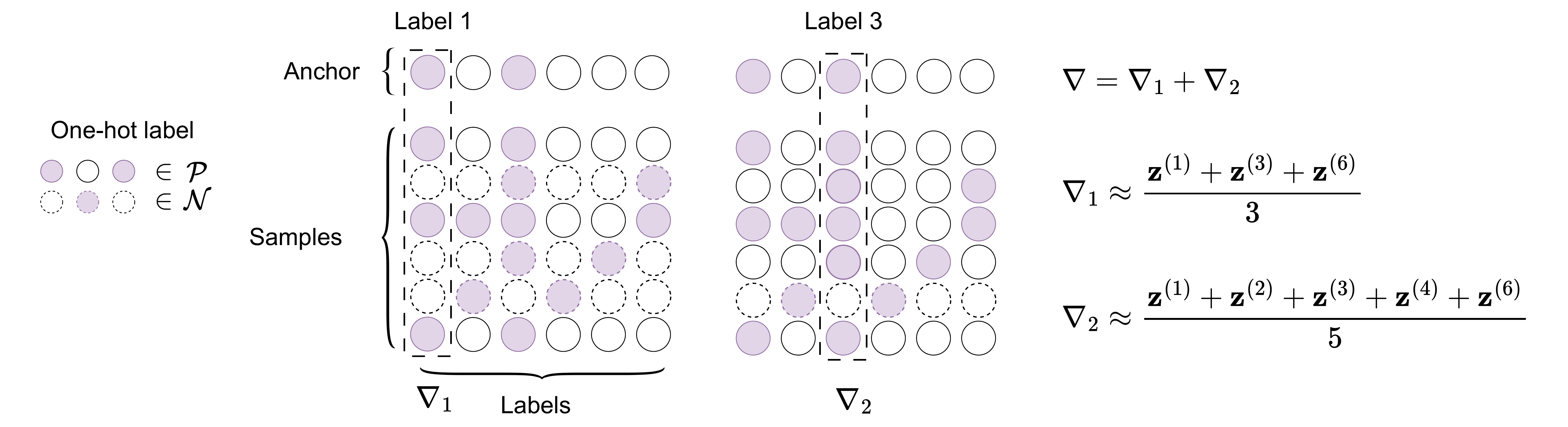}
\caption{
Illustration of the MulSupCon method.
Each row represents a sample’s one-hot label vector, with the first row 
being the anchor sample (top row) and the remaining rows corresponding to other 
samples in the batch. Filled circles indicate active label dimensions.
Samples enclosed with dotted outlines belong to the negative set 
$\mathcal{N}$, while the remaining samples form the positive set 
$\mathcal{P}$. For an anchor with labels $\{1,3\}$, the corresponding 
positive sets are $P_1 = \{1,3,6\}$ and $P_3 = \{1,2,3,4,6\}$, which form 
the basis for label-wise contrastive learning in MulSupCon.
}
\label{fig:mulsupcon}
\end{center}
\end{figure*}

\subsection{Multi-Label Remote Sensing Image Retrieval (MLRSIR)}
MLRSIR addresses the content-based retrieval of satellite scenes with multiple co-occurring semantic elements \cite{imbriaco2022toward}. Each image typically contains several land cover categories such as residential areas, roads, vegetation, or water, which have appeared at different spatial scales and formed highly heterogeneous scene compositions \cite{Zhou2020Region}, that complicate learning accurate representations reflecting image similarity \cite{sumbul2022informative}. Furthermore, the distribution of land cover classes is often imbalanced, thus hindering the learning of balanced retrieval embeddings \cite{Fu2024GenerativeCL}.

Notable approaches to MLRSIR include graph based frameworks that have been proposed to model label dependencies and spatial relationships \cite{Chaudhuri2018IEEE, Sumbul2021IGARSS, Kang2021GraphRelationNetwork}, while reconstruction and segmentation driven pipelines have performed retrieval using region or pixel level masks \cite{dai2017novel}. Region convolutional and fully convolutional feature extraction schemes have also been developed to enhance spatially localized retrieval performance \cite{ shao2020multilabel}. Moreover, attention based and transformer based architectures have also been investigated to capture contextual information and long range dependencies among labels \cite{Hua2021}. 

Despite the progress achieved by these architectural approaches, their performance remains constrained by the quality of the underlying feature space. As a result, increased attention has been directed toward multi-label contrastive learning, which offers a principled mechanism for representing semantic overlap and inter-label relationships more effectively.

\subsection{Multi-label contrastive learning}

Multi-label contrastive learning was initially developed within computer vision, and its principles have been considered applicable to MLRSIR, where images are characterized by multiple co-occurring categories, imbalanced label distributions, and complex label relationships. Earlier MLRSIR research has been primarily directed toward architectural designs, while contrastive and metric-learning formulations have been recognized as a strong alternative for improving multi-label feature representations. Their general and flexible objectives, originally developed for multi-label image understanding, have been adapted to remote sensing image retrieval, particularly in scenarios defined by diverse label co-occurrence patterns and long-tailed distributions.

A large portion of existing approaches are based on the foundational Supervised Contrastive Learning (SupCon) objective \cite{khosla2021supcon}, which encourages semantically similar samples to be grouped in embedding space. Because SupCon was designed for single-label supervision, multi-label variants have mainly differed in the definition of positive samples and in the weighting of label relationships.
Early extensions follow two simple strategies: SupCon–ALL, which treats samples as positives only when their complete label sets match, and SupCon–ANY, which treats samples as positives when they share at least one label. These strategies represent strict and inclusive supervision, but they do not reflect differences in the strength of partial semantic overlap.
More refined approaches have therefore been developed. Soft-weighting schemes such as LBase \cite{audibert2024exploring} use Jaccard similarity to assign continuous weights according to label-set overlap. Prototype-based methods such as LProto  \cite{gupta2023class} shift learning toward instance–prototype alignment, producing more stable class-level representations. Label-wise strategies such as MulSupCon \cite{zhang2024mulsupcon} and its weighted variant treat each label of a sample as a separate anchor, providing finer supervision and more balanced gradients for samples with many labels.
Beyond simple overlap metrics, some formulations incorporate structural or distributional properties of label space. LMSC \cite{audibert2024exploring}uses prototypes, memory queues, and frequency-aware reweighting to address long-tailed label distributions. LReg \cite{audibert2024survey}introduces gradient regularization to reduce conflicts among correlated labels, while SimDis \cite{huang2024similarity} models both similarity and dissimilarity through label intersections and complements to provide finer semantic control.

Despite their differences, these methods rely on two central design choices: how positive samples are defined and how label-dependent weighting influences attraction and repulsion during contrastive learning.
However, most existing methods still depend on static assumptions regarding label relations, typically expressed through overlap counts, binary relevance matrices, or predefined graphs. This static treatment limits the ability of these models to handle label imbalance, semantic variability, and uneven co-occurrence frequencies, all of which are pronounced in remote sensing imagery.

These observations have motivated the development of a more adaptive contrastive formulation that can adjust similarity forces according to label rarity, semantic consistency, and sample complexity. In response, the Multi-Label Adaptive Contrastive Learning (MARC) framework is introduced, in which adaptive weighting and dynamic temperature scaling are used to better align multi-label relationships and improve retrieval performance in remote sensing scenarios.

\section{MARC}
\label{sec:proposed_loss}

The MARC framework is designed to overcome the limitations of existing multi-label contrastive methods which rely on uniform pair weighting and fixed temperature scaling. MARC introduces two adaptive mechanisms, Pairwise Label Reweighting (PLR) and Dynamic Temperature Scaling (DTS), allowing the loss to incorporate semantic imbalance, label co-occurrence, and inter-class relationship awareness within contrastive learning.

\begin{figure*}[!t]
\begin{center}
\includegraphics[width=\linewidth,height=0.48\textheight,keepaspectratio]
{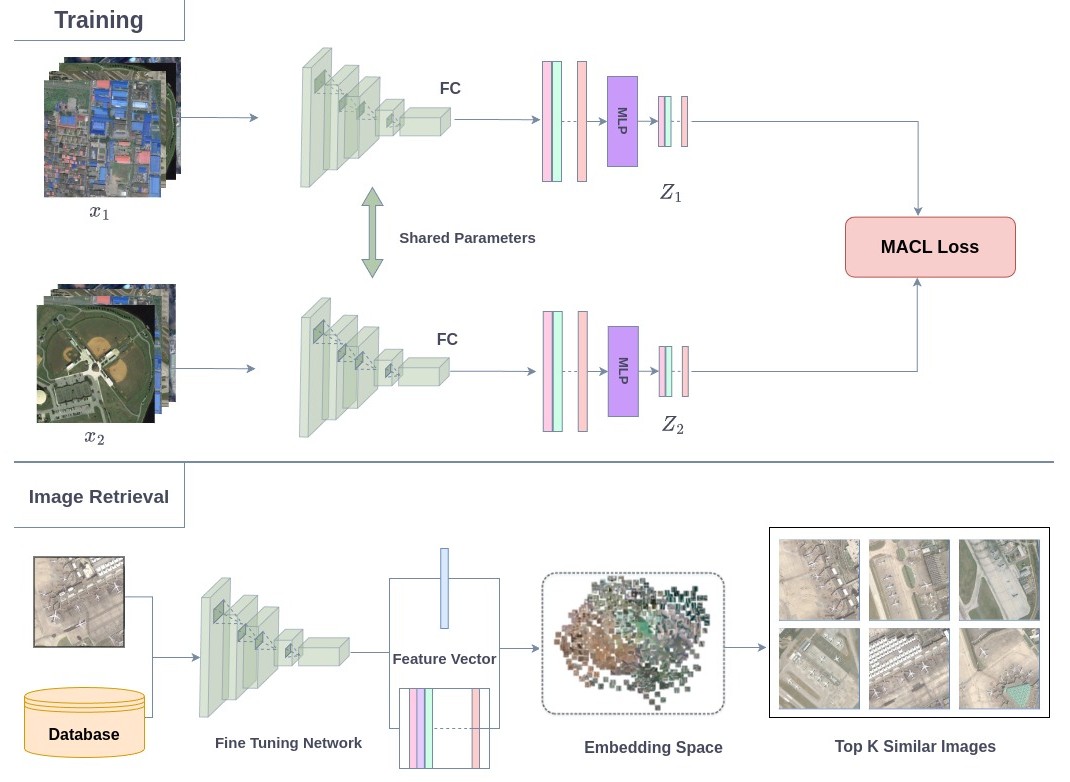}

\caption{%
Training and retrieval pipeline employing the proposed MARC loss.
During training, images are encoded by a shared backbone and projection head to
produce embeddings optimized using the MARC loss.
During retrieval, query and database images are embedded into a common feature
space and ranked based on similarity.}
\label{fig:MARC_architecture}
\end{center}
\end{figure*}

\textbf{Motivation:} MLRSIR presents unique challenges: labels are overlapping, imbalanced, and strongly correlated. Standard SupCon is inherently limited to single-label settings. MulSupCon extends SupCon to multi-label data by decomposing each anchor into multiple label-specific anchors and treating samples sharing that label as positives, but it still relies on simplifying assumptions. To visually demonstrate how MulSupCon constructs label-wise positive sets and applies uniform contributions, Fig.~\ref{fig:mulsupcon} is presented as a schematic illustration, which also facilitates a smoother transition into the limitations discussed next.

These assumptions are problematic in remote sensing because frequent classes (e.g., \emph{building}, \emph{road}) dominate gradients, rare but informative classes (e.g., \emph{airplane}, \emph{harbor}) are underrepresented, and semantic proximity varies widely (e.g., \emph{residential} $\leftrightarrow$ \emph{road} vs. \emph{harbor} $\leftrightarrow$ \emph{stadium}).

Case 1 (Unequal semantic overlap): An anchor labeled $\{building, road, tree\}$ compared with a sample labeled $\{building\}$ should be considered less similar to it than when compared with a sample labeled $\{building, road, tree\}$, yet conventional losses treat both pairs equally.

Case 2 (Rare vs. frequent classes): Pairs involving common labels (e.g., \emph{pavement}) dominate training, while rare but critical labels (e.g., \emph{airplane}) contribute little, biasing embeddings in favor of frequent classes.

The introduction of MARC is motivated by the fact that uniform pair treatment fails to capture semantic imbalance, rarity, and heterogeneous label interactions.
MARC is formulated as a generalization that replaces uniform pair treatment with adaptive, relationship-aware optimization.
Positive pairs within each label-specific set vary in importance and semantic strength, and their contributions are determined by label rarity, semantic overlap, and co-occurrence patterns, thereby aligning optimization with the intrinsic structure of multi-label data.

\paragraph{MARC Architecture and Workflow.}
As illustrated in Fig.~\ref{fig:MARC_architecture}, MARC follows an embedding-based
contrastive learning paradigm. During training,
images within a mini-batch are processed by a shared backbone encoder followed by
a projection head, producing normalized feature embeddings in a shared
representation space. For each anchor image, all remaining samples in the batch
act as comparisons, forming label-dependent positive and negative sets according
to their multi-label annotations. The proposed MARC loss is then applied to these
embeddings, adaptively modulating pairwise interactions through PLR and DTS.

At inference time, the trained encoder is used to embed both query and database
images into the same feature space. Image retrieval is performed by ranking
database embeddings based on similarity to the query embedding, without requiring
label information. This unified architecture ensures that the adaptive contrastive
optimization learned during training directly translates to improved retrieval
performance under multi-label supervision.

\begin{figure*}[!t]
\begin{center}
    \includegraphics[width=\linewidth]{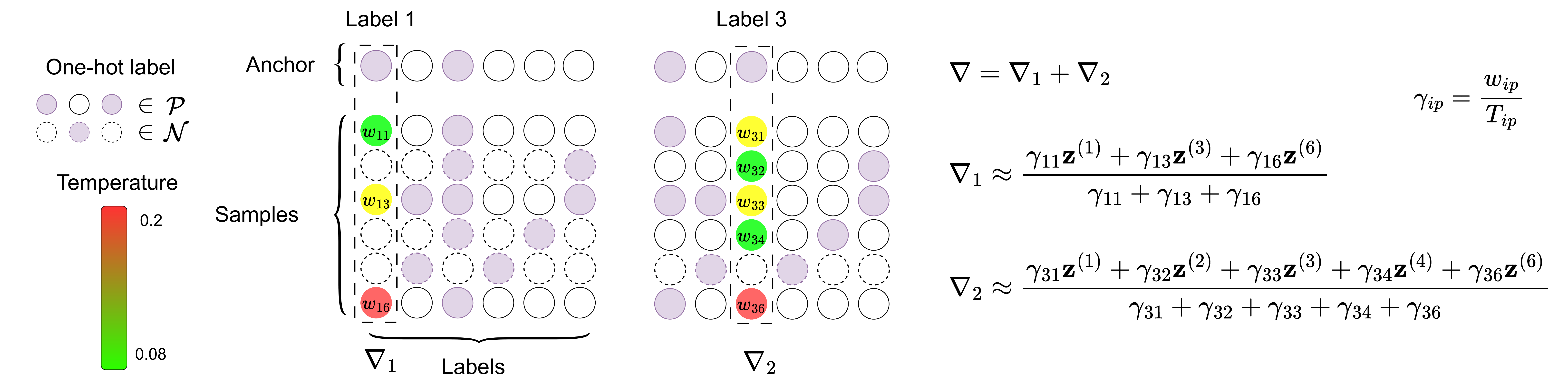}
    \caption{
Visualization of the MARC mechanism. For each anchor label, MARC computes a label-specific gradient component using pairwise weights (green/yellow/red nodes) and adaptive temperatures (color bar). The effective influence of each positive is determined by $\gamma_{ip} = w_{ip}/T_{ip}$, and the final gradient is obtained by combining the label-specific components.}
\label{fig:MARC}
\end{center}
\end{figure*}

In contrast to traditional contrastive methods, which treat all positive pairs uniformly, MARC treats each pair as a unique learning relationship with distinct importance and difficulty. 
This is accomplished through two synergistic mechanisms: PLR and DTS.

Before the mechanisms are introduced, the notation used throughout the formulation is summarized for clarity.

\textbf{Notation: }
The embedding of the anchor sample is written as $\mathbf{z}_q^{(i)}$, and the embedding of any comparison is $\mathbf{z}_k^{(a)}$. The pairwise similarity is expressed as $s_{ia} = \mathbf{z}_q^{(i)} \cdot \mathbf{z}_k^{(a)}$, and the set of samples compared with anchor $i$ is represented by $A(i)$.The label set associated with sample $i$ is denoted by $y^{(i)}$, and the number of labels assigned to that sample is written as $|y^{(i)}|$. A specific label within this set is referred to as $y_j^{(i)}$, indicating the $j$th label of sample $i$.

\paragraph{PLR:} It establishes a \emph{rarity-aware importance hierarchy} by assigning each anchor–positive pair $(i,p)$ a weight $w_{ip}$ that reflects the informativeness of their shared labels. The pairwise weight $w_{ip}$ quantifies how rare the shared labels are between anchor $i$ and positive sample $p$ and is computed using the label-intersection frequency function $f(y^{(i)}, y^{(p)})$, which measures how often the two label sets co-occur in the dataset. Rare co-occurrences lead to larger weights:

\begin{equation}
    w_{ip} = \frac{1}{\log(1 + f(y^{(i)}, y^{(p)})) + \epsilon}
\end{equation}

This weighting ensures that informative, low-frequency relationships exert stronger influence than common label pairs.

\paragraph{DTS:} It introduces \emph{semantic-aware similarity calibration} by recognizing that different label combinations require different similarity thresholds. The pairwise temperature $T_{ip}$ adapts the similarity calibration based on two components: the semantic-overlap measure $J(y^{(i)}, y^{(p)})$, which quantifies how similar the two label sets are, and the marginal frequency $h(y^{(i)})$, which represents how rare anchor $i$'s labels are in the dataset. 
This relationship is formally defined as:

\begin{equation}
T_{ip} =
\underbrace{
\exp(-\alpha J(y^{(i)}, y^{(p)}))
}_{\text{Term 1: Pair Overlap}\\[4pt]}
+
\underbrace{
\beta \left( \frac{1}{\log(1 + h(y^{(i)}))} \right)
}_{\text{Term 2: Anchor Rarity}\\[4pt]}
\end{equation}

High semantic overlap results in lower temperatures, while rare anchors receive additional temperature increases to stabilize optimization. This adaptive temperature encourages sharper boundaries for semantically close pairs and smoother ones for weakly related pairs.

Here, $J(\cdot)$ denotes the semantic-overlap measure, $h(\cdot)$ denotes the marginal label frequency, and $\alpha$ and $\beta$ control semantic emphasis and rarity weighting, respectively.
Together, PLR and DTS establish a heterogeneous similarity landscape in which each anchor--positive pair contributes according to its semantic strength, rarity, and contextual informativeness.

As shown in Fig.~\ref{fig:MARC}, MARC assigns each positive a weighted and temperature-scaled contribution, determining how strongly that sample influences the anchor. 
This produces heterogeneous, label-aware gradients that reflect semantic overlap, rarity, and pairwise informativeness.

The family of supervised contrastive learning methods is generalized by MARC. Classical methods such as SupCon, MulSupCon, and LBase appear as special cases when PLR weights are set to $w_{ip}=1$ and DTS temperatures are fixed to a constant $\tau$.
Through adaptive weighting and temperature scaling, MARC expands the contrastive optimization space beyond uniform objectives.

This expanded flexibility is reflected in several ways:

\begin{itemize}
    \item Each pair’s contribution is determined by semantic relatedness and label rarity, producing a gradient field aligned with multi-label structure.
    \item Gradient flow varies across the embedding space based on pair difficulty instead of uniformly across all positives.
    \item Convergence is shaped jointly by global label statistics and local semantic interactions.
\end{itemize}

\paragraph*{MARC loss Formulation:} MARC extends MulSupCon by incorporating its two adaptive mechanisms (PLR and DTS) into each label-wise contrastive component.
For each $y_j^{(i)} \in \mathbf{y}^{(i)}$, a separate positive set has been constructed:

\begin{equation}
    P^{(i)}_j=\{\,m \mid \forall m,\; y^{(i)}_j \in y^{(m)}\,\}
\end{equation}

For anchor $i$ with label set $\mathbf{y}^{(i)}$, MARC loss is defined as:

\begin{equation}
\begin{split}
\mathcal{L}_{\text{MARC}}
= \sum_{y^{(i)}_j \in y^{(i)}} 
  \frac{-1}{|P^{(i)}_j|}
  \sum_{p \in P^{(i)}_j}
  w_{ip}\,
  \log
  \frac{
    \exp(s^{(i)}_p / T_{i,p})
  }{
    \sum_{a \in A^{(i)}} 
    \exp(s^{(i)}_a / T_{i,a})
  }
\end{split}
\end{equation}

Here, $w_{ip}$ and $T_{ip}$ are the PLR and DTS terms, respectively, determining the adaptive strength and weighting of each positive relationship.
Fig.~\ref{fig:MARC} visualizes how these pair-specific terms produce non-uniform contributions across label-dependent positive sets.

MARC discovers and exploits the inherent heterogeneity of multi-label relationships. Contrastive learning from a rigid, uniform process into a flexible, adaptive optimization that respects the complexity of real-world multi-label data.
From a geometric perspective MARC creates a dynamic embedding space where the interaction structure of attraction and repulsion is continuously modulated based on relationship properties. The embedding space exhibits adaptive spatial structure where important relationships create stronger gradients and semantically rich pairs form tighter clusters.

\paragraph*{\textbf{Gradient Analysis of MARC:}}
The behavior of MARC in the embedding space can be examined by analyzing the gradient of the loss with respect to the anchor embedding $\mathbf{z}_q^{(i)}$.
The MARC objective is given by

\begin{equation}
\mathcal{L}_{\text{MARC}}^{(i)}
=
\sum_{y_j^{(i)} \in y^{(i)}}
\frac{-1}{|P_j^{(i)}|}
\sum_{p \in P_j^{(i)}} 
w_{ip}\,
\log
\frac{\exp(s_{ip}/T_{ip})}
{\sum_{a \in A(i)} \exp(s_{ia}/T_{ia})}
\end{equation}

where 
\begin{equation}
s_{ia} = \mathbf{z}_q^{(i)} \cdot \mathbf{z}_k^{(a)}
\end{equation}

Substituting this relation yields the gradient expression:

\begin{equation}
\begin{split}
\nabla_{\mathbf{z}_q^{(i)}} \mathcal{L}_{\text{MARC}}^{(i)}
=
&\sum_{y_j^{(i)} \in y^{(i)}}
\left[
-\frac{1}{|P_j^{(i)}|}
\sum_{p\in P_j^{(i)}}
\frac{w_{ip}}{T_{ip}}\, \mathbf{z}_k^{(p)}
\right]
\\
&+
|y^{(i)}|
\left[
\sum_{a\in A(i)}
\frac{1}{T_{ia}}
\frac{\exp(s_{ia}/T_{ia})}{C}\,
\mathbf{z}_k^{(a)}
\right]
\end{split}
\end{equation}

The adaptive attraction term for label $y_j^{(i)}$ is defined as:

\begin{equation}
\bar{\mathbf{z}}_j^{\text{MARC}}
=
-\frac{1}{|P_j^{(i)}|}
\sum_{p\in P_j^{(i)}}
\frac{w_{ip}}{T_{ip}}\, \mathbf{z}_k^{(p)}
\end{equation}

while the adaptive repulsion term is expressed as:

\begin{equation}
\hat{\mathbf{z}}^{\text{MARC}}
=
\sum_{a\in A(i)}
\frac{1}{T_{ia}}
\frac{\exp(s_{ia}/T_{ia})}{C}\,
\mathbf{z}_k^{(a)}
\end{equation}

Combining these components yields the compact gradient form

\begin{equation}
\nabla_{\mathbf{z}_q^{(i)}} \mathcal{L}_{\text{MARC}}
=
\sum_{y_j^{(i)} \in y^{(i)}} \bar{\mathbf{z}}_j^{\text{MARC}}
+
|y^{(i)}| \cdot \hat{\mathbf{z}}^{\text{MARC}}
\end{equation}

The vector $\bar{\mathbf{z}}^{\text{MARC}}_j$ acts as a label-specific adaptive representation, aggregating positive samples in proportion to their rarity and semantic closeness. The repulsion vector $\hat{\mathbf{z}}^{\text{MARC}}$ prevents collapse by distributing temperature-scaled force across all comparisons.

For any sample $t$ sharing the label intersection $y = y^{(i)} \cap y^{(t)}$, the total attraction is:
\[
-\sum_{y_j \in y}
\frac{w_{it}}{T_{it}}
\frac{1}{|P^{(i)}_j|}
\]

This shows that samples with greater semantic strength exert stronger pull, while common or weakly related samples contribute less influence.

\textbf{Complexity:} 
MARC introduces pair-specific weights and temperatures, resulting in $\mathcal{O}(B^2)$ computational complexity for batch size $B$. 
Label co-occurrence statistics and overlap computations are precomputed and cached, ensuring minimal runtime overhead.
In practice, the added computations are lightweight and do not significantly impact runtime.

Overall, MARC transforms multi-label contrastive learning from a uniform interaction model into a fully adaptive, relationship-aware framework.
By tailoring gradients to semantic relevance, rarity, and label overlap, MARC produces a structured embedding space that aligns with the heterogeneous nature of real-world multi-label data.

\begin{table*}[!t]
\begin{center}
\caption{Summary of selected MLRSIR datasets.}
\label{tab:mlrsir_datasets_summary}
\renewcommand{\arraystretch}{1.2}
\begin{tabular}{c|c|c|c|c|c|c}
\hline
\textbf{Dataset} & \textbf{No. of Labels} & \textbf{No. of Images} & \textbf{Image Size} & \textbf{Resolution (m)} & \textbf{Image Source} & \textbf{Year} \\
\hline
DLRSD\cite{chaudhuri2018multilabel} & 17 & 2,100 & 256$\times$256 & 0.3   & DLRSD                   & 2018 \\
ML-AID\cite{hua2020relation}       & 17 & 3,000 & 600$\times$600 & 8--0.5 & AID                     & 2020 \\
WHDLD\cite{shao2020multilabel}     & 6  & 4,940 & 256$\times$256 & 2     & Gaofen-1 and Ziyuan-3   & 2020 \\
\hline
\end{tabular}
\end{center}
\end{table*}

\section{Experiments}
The experiments have been conducted with the goal of evaluating MARC against existing supervised contrastive losses on standard ML-RSIR benchmarks (Tables \ref{tab:mlrsir_datasets_summary}).

\subsection{Datasets and splitting}

\textbf{DLRSD} \cite{chaudhuri2018multilabel}: provides aerial images annotated with binary multi-hot vectors across 21 land-use categories. It is widely used in multi-label remote sensing because it captures diverse urban and natural scenes at high spatial resolution.

\textbf{ML-AID} \cite{hua2020relation}: extends the AID dataset with multi-label annotations, enabling evaluation of both classification and retrieval tasks. Each image is annotated with multiple land-cover classes, and labels are provided in CSV format.

\textbf{WHDLD} \cite{shao2020multilabel}: is a dense pixel-level labeling dataset derived from Gaofen-1 and Ziyuan-3 imagery. Unlike DLRSD and ML-AID, it provides annotations for six primitive land-cover categories, making it suitable for both image-level and pixel-level analysis.

\begin{table}[!h]
\caption{Overview of dataset train/val/test splits.}
\label{tab:datasets}
\begin{center}
\begin{tabular}{l|c|c|c|c|c}
  \toprule
  \textbf{Dataset} & \textbf{\#Images} & \textbf{\#Labels} & \textbf{Train} & \textbf{Validation} & \textbf{Test} \\
  \midrule
  DLRSD    & 2,100    & 17       & 1,475    & 214      & 411   \\
  ML-AID   & 3,000    & 17       & 2,100    & 300      & 600    \\
  WHDLD    & 4,940    & 6        & 3,444    & 518      & 978   \\
  \bottomrule
\end{tabular}
\end{center}
\end{table}

\begin{table*}[!t]
\begin{center}
\caption{Overview of the DLRSD, ML-AID, and WHDLD datasets, including class names and the no. of images per label.}
\label{tab:mlrsir_datasets}

\begin{tabular}{c|c|c|c|c|c|c}
\hline
\textbf{Label} &
\textbf{DLRSD Class} &
\textbf{No. of Images} &
\textbf{ML-AID Class} &
\textbf{No. of Images} &
\textbf{WHDLD Class} &
\textbf{No. of Images} \\ [6 pt]
\hline
1  & airplane     & 100  & bare soil   & 1475 & building   & 3722 \\
2  & bare soil    & 754  & airplane    & 99   & road       & 3162 \\
3  & building     & 713  & building    & 2161 & pavement   & 3881 \\
4  & car          & 897  & car         & 2026 & vegetation & 4631 \\
5  & chaparral    & 116  & chaparral   & 112  & bare soil  & 3539 \\
6  & court        & 105  & court       & 344  & water      & 3886 \\
7  & dock         & 100  & dock        & 271  & --         & --   \\
8  & field        & 103  & field       & 214  & --         & --   \\
9  & grass        & 977  & grass       & 2295 & --         & --   \\
10 & mobile home  & 102  & mobile home & 2    & --         & --   \\
11 & pavement     & 1331 & pavement    & 2328 & --         & --   \\
12 & sand         & 291  & sand        & 259  & --         & --   \\
13 & sea          & 101  & sea         & 221  & --         & --   \\
14 & ship         & 103  & ship        & 284  & --         & --   \\
15 & tank         & 100  & tank        & 108  & --         & --   \\
16 & tree         & 1021 & tree        & 2406 & --         & --   \\
17 & water        & 208  & water       & 852  & --         & --   \\
\hline
\end{tabular}
\end{center}
\end{table*}

Following the approach in\cite{imbriaco2022toward}, all datasets are partitioned using a fixed 70/10/20 ratio for training, validation, and testing, respectively. The splits are generated randomly for each dataset (Table \ref{tab:datasets}). In the retrieval stage, each test image is used once as a query, and the remaining test images form the gallery. This setup ensures consistent and fair evaluation across datasets with different sizes, label distributions, and spatial resolutions.

\subsection{Implementation Details}
\begin{figure}[!t]
\begin{center}

    \subfloat[DLRSD example 1]{
        \includegraphics[width=0.475\textwidth,height=0.5\textheight,keepaspectratio]{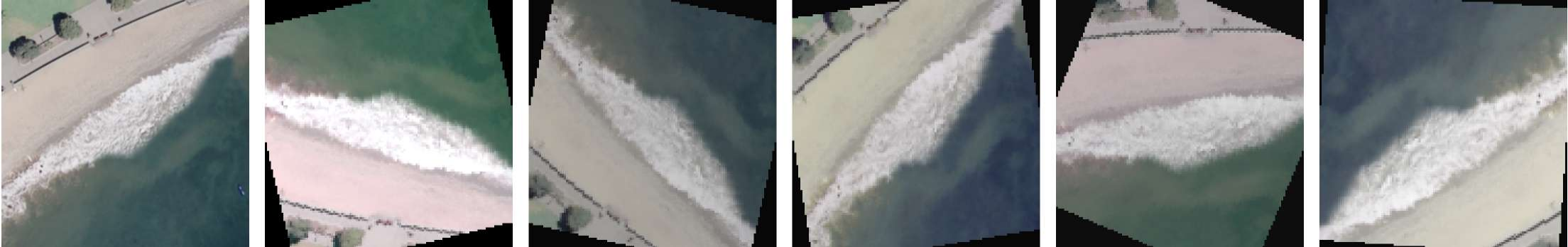}
        \label{fig:dlrsd_a}
    }    
    \subfloat[DLRSD example 2]{
        \includegraphics[width=0.475\textwidth,height=0.5\textheight,keepaspectratio]{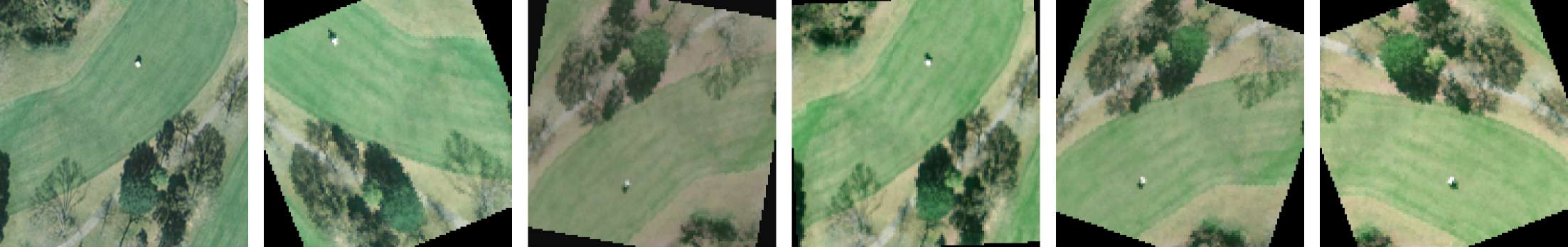}
        \label{fig:dlrsd_b}
    }

\end{center}
\caption{Examples of data augmentations applied to the DLRSD dataset.}
\label{fig:dlrsd_augmentations}
\end{figure}

\begin{figure}[!t]
\begin{center}

    \subfloat[ML-AID example 1]{
        \includegraphics[width=0.475\textwidth,height=0.8\textheight,keepaspectratio]{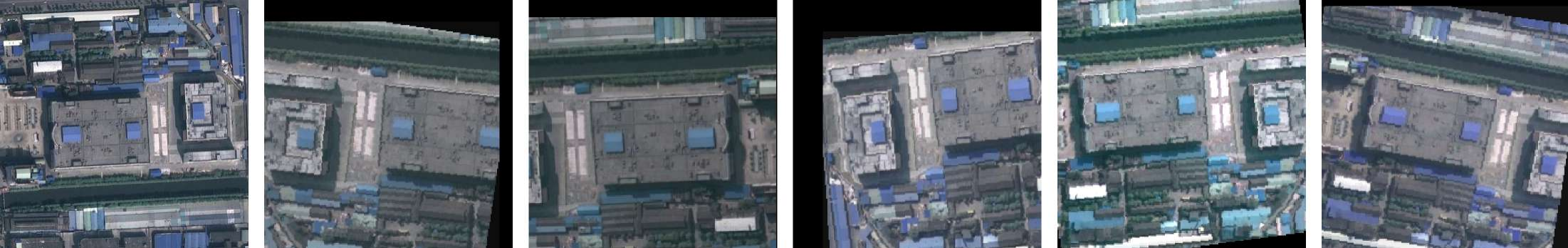}
        \label{fig:mlaid_a}
    }%
    \subfloat[ML-AID example 2]{
        \includegraphics[width=0.475\textwidth,height=0.8\textheight,keepaspectratio]{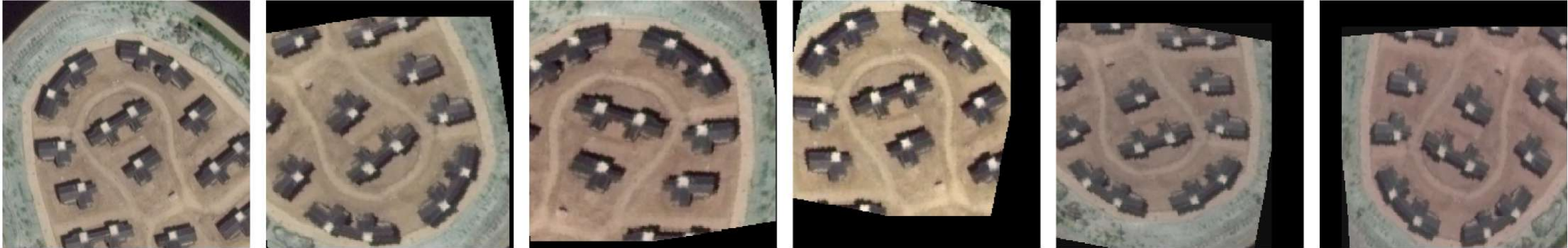}
        \label{fig:mlaid_b}
    }

\end{center}
\caption{Examples of data augmentations applied to the ML-AID dataset.}
\label{fig:mlaid_augmentations}
\end{figure}

\begin{figure}[!t]
\begin{center}

    \subfloat[WHDLD example 1]{
        \includegraphics[width=0.475\textwidth,height=0.8\textheight,keepaspectratio]{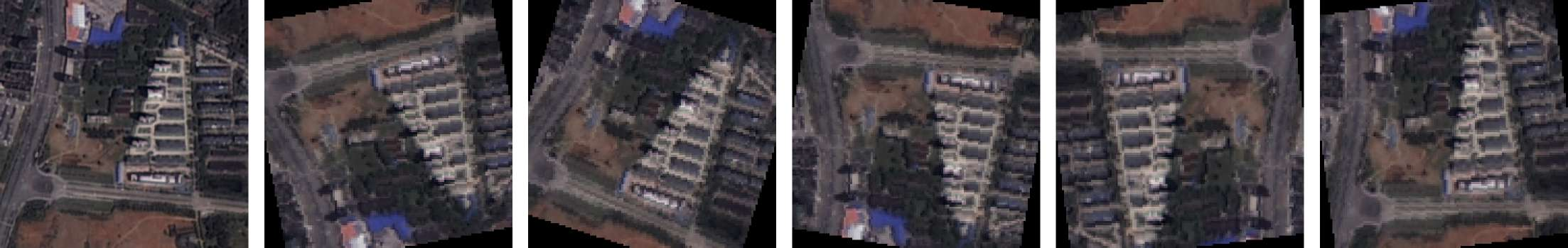}
        \label{fig:whdld_a}
    }
    \subfloat[WHDLD example 2]{
        \includegraphics[width=0.475\textwidth,height=0.8\textheight,keepaspectratio]{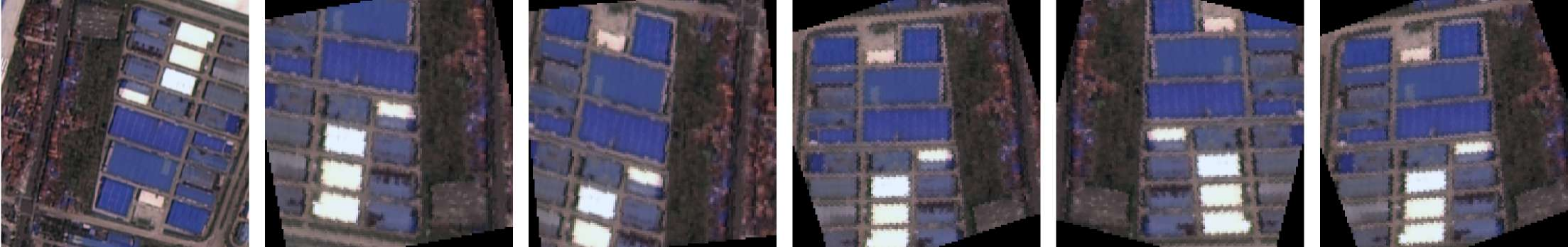}
        \label{fig:whdld_b}
    }

\end{center}
\caption{Examples of data augmentations applied to the WHDLD dataset.}
\label{fig:whdld_augmentations}
\end{figure}

The framework was developed using ResNet-18 as the backbone and a two-layer projection head. To enhance the diversity of training data and improve model robustness, several augmentations were applied, including random resized cropping, horizontal and vertical flipping, small-angle rotations, and color jittering. Examples of augmented samples for each dataset are shown in Figs.~\ref{fig:dlrsd_augmentations}, \ref{fig:mlaid_augmentations}, and \ref{fig:whdld_augmentations}. In practice, MARC has training and inference costs comparable to MulSupCon, since PLR and DTS only introduce lightweight pairwise computations without altering backbone architecture or feature dimensionality.

Model training was conducted on an NVIDIA GeForce GTX 1660 Ti GPU with 6\,GB of memory. Each model underwent 100 epochs of training. The process used the Adam optimizer (learning rate of 0.001, temperature of 0.3, weight decay of 0.0005) with a batch size of 128. Additionally, a CosineAnnealingLR scheduler lowered the learning rate by 20\% every 15 epochs, and gradient clipping (with a maximum norm of 1.0) was applied to prevent exploding gradients. For numerical stability, The value $\epsilon = 10^{-8}$ was set in PLR to avoid division by zero. In DTS, $\alpha = 1.5$ was set to control the influence of semantic similarity, and $\beta = 0.1$ was set to amplify the effect of rare labels. All hyperparameters were selected empirically based on validation performance and stability considerations.

\subsection{Evaluation Metrics}
Retrieval performance is evaluated using two categories of metrics: cosine-similarity-based metrics for ranking quality in the learned embedding space, and Jaccard-similarity-based metrics for multi-label relevance assessment.

\subsubsection{\textbf{Cosine-Similarity Based Metrics}}
Cosine-similarity-based metrics evaluate ranking quality in the learned embedding space by comparing feature vectors of query and database images. Two points are considered similar if they belong to the same category or share at least one tag. Following\cite{cao2023deep_continual_hashing}, mAP@5000 computes the mean average precision over the top-5000 retrieved results. The nDCG@100 assesses the top-100 ranking using graded relevance, with $\mathrm{IDCG}@K$ denoting the ideal ordering for $k = 100$.

\begin{equation}
\begin{split}
P@n &= \frac{\#\{\text{relevant images in top $N$ results}\}}{N} \\[6pt]
AP\phantom{n} &= \frac{\sum_{n} P@n \times I\{\text{image $n$ is relevant}\}}
          {\#\{\text{retrieved relevant images}\}} \\[6pt]
\mathrm{mAP} &= \frac{1}{Q} \sum_{i} AP_i
\end{split}
\end{equation}

where $\#$ indicates a count function, $I$ denotes an indicator function, and $Q$ is the total number of queries. The mAP is computed from the retrieval list over the entire database, whereas mAP@$n$ is calculated from only the top-$n$ retrieved results.

\begin{equation}
\mathrm{nDCG}@p = \frac{1}{Z} \sum_{i=1}^{p} \frac{2^{\,r_i} - 1}{\log(1 + i)}
\end{equation}

\noindent where $Z$ represents the ideal $\mathrm{DCG}@p$, computed from the correct ranking list. Here, $r(i) = |\mathbf{y}_q \cap \mathbf{y}_i|$ indicates the similarity between the $i$-th item and the query.


\subsubsection{\textbf{Jaccard-Similarity Based Metrics}}
Jaccard-based metrics assess semantic similarity between query and database images using the Jaccard similarity matrix, ranking results by label overlap. Following the multi-label retrieval protocol in\cite{imbriaco2022toward}, three measures are used: mean Average Precision (mAP), normalized Discounted Cumulative Gain (nDCG), and weighted Average Precision (wAP). The mAP is computed under three thresholds \textbf{(Easy: 0.40, Medium: 0.60, Hard: 0.80)} to capture different levels of label overlap. The nDCG measures ranking quality using graded Jaccard relevance and is reported at $k = 100$. The wAP measures global precision weighted by the number of shared labels, without requiring a fixed relevance threshold, also at $k = 100$.

\paragraph{\textbf{Mean Average Precision (mAP)}}
Let $Q$ be the number of queries. The mAP is given by:
\begin{equation}
\mathrm{mAP} = \frac{1}{Q} \sum_{q}^{Q} \mathrm{AP}(q),
\end{equation}
where AP the average precision for a query $q$ is:
\begin{equation}
\mathrm{AP}(q) = \frac{1}{N_{\phi}(q)@k} \sum_{i}^{k} \ \phi(q,i) \frac{N_{\phi}(q)@i}{i}
\end{equation}

\noindent where $\phi(q, i)$ is the indicator function for query $q$ and the $i$-th image, and $N_{\phi}(q)@k$ is the number of positive images in the top-$k$ ranks. Three settings for $\phi$ have been used to assess image matching difficulty: \emph{Easy}, \emph{Medium}, and \emph{Hard}, with Jaccard Index thresholds of $0.40$, $0.60$, and $0.80$, respectively. These protocols evaluate performance under varying label counts, enable intuitive result interpretation, and emphasize label similarity in ML-RSIR and correct ranking (higher Jaccard Index ranked higher).

\paragraph{\textbf{Normalized Discounted Cumulative Gain (nDCG)}}
The nDCG is suitable for multilabel retrieval tasks, as it evaluates the relevance of the top-ranked results. It is defined at $k$ as
\begin{equation}
\mathrm{nDCG}@k = \frac{\mathrm{DCG}@k}{\mathrm{IDCG}@k}
\end{equation}
where the $\mathrm{DCG}@k$ for a query $q$ is
\begin{equation}
\mathrm{DCG}@k = \sum_{i=1}^{k} \frac{2^{\,J(q,i)} - 1}{\log_2(i+1)}
\end{equation}
with $J(q,i)$ denoting the Jaccard Index between the query and the $i$-th ranked image. The ideal discounted cumulative gain ($\mathrm{IDCG}@k$) is computed as
\begin{equation}
\mathrm{IDCG}@k = \sum_{i=1}^{k} \frac{2^{\,J(q,i)_{r}} - 1}{\log_2(i+1)}
\end{equation}
where $J(q,i)_{r}$ is the Jaccard Index between the query and the $i$-th ranked image, sorted in decreasing order of relevance $r$. The nDCG value ranges from zero to one, making results and comparisons more interpretable.

\paragraph{\textbf{Weighted Average Precision (wAP)}}
The weighted Average Precision (wAP) metric is analogous to mAP, but incorporates the number of shared classes between the query and each retrieved image in its computation. This makes it applicable to ML-RSIR without requiring a predefined threshold for the indicator function. The wAP is therefore calculated as

\begin{equation}
\mathrm{wAP} = \frac{1}{Q} \sum_{q}^{Q} \frac{1}{N_{\phi}(q)@k} 
\sum_{i}^{k} \phi(q,i) \left( \frac{1}{i} \sum_{j}^{i} J(q,j) \right)
\end{equation}

\section{Results and Discussion}

\subsection{Quantitative Results}

The quantitative evaluation is organised per dataset. Retrieval performance is 
reported using cosine similarity–based metrics and relevance-based Jaccard 
measures in order to capture both global embedding quality and 
threshold-dependent retrieval behaviour.


\begin{table*}[!t]
\begin{center} 
\caption{Retrieval results for DLRSD where \textcolor{red}{red} and \textcolor{blue}{blue} indicate the highest and second highest values respectively.}
\label{tab:results_dlrsd}  
\fontsize{8pt}{12pt}\selectfont
\begin{tabular}{c|cc|ccc|cc}
\hline
\multirow{3}{*}{\textbf{Method}} & \multicolumn{2}{c|}{\multirow{1}{*}{\textbf{Cosine Similarity-Based}}} & \multicolumn{5}{c}{\multirow{1}{*}{\textbf{Relevance-Based Jaccard Similarity}}} \\
\cline{2-8}
 & \multirow{2}{*}{\textbf{mAP(sim)@5000}} & \multirow{2}{*}{\textbf{nDCG(sim)@100}} & \multicolumn{3}{c|}{\textbf{mAP Threshold}} & \multirow{2}{*}{\textbf{nDCG@100}} & \multirow{2}{*}{\textbf{wAP@100}} \\
 
\cline{4-6}
 &  &  & \textbf{Easy@0.4} & \textbf{Medium@0.6} & \multicolumn{1}{c|}{\textbf{Hard@0.8}} & &  \\
\hline 
SupCon ALL \cite{khosla2021supcon}            & 68.72 & 71.14 & 30.12 & 13.45 & 5.32 & 73.27 & 14.30 \\
SupCon ANY \cite{khosla2021supcon}           & 68.43 & 69.85 & 30.13 & 13.55 & 5.49 & 72.42 & 13.76 \\
LBase   \cite{audibert2024exploring}         & 69.14 & 71.43 & \textbf{\textcolor{red}{30.36}} & 13.47 & 5.12 & 74.00 & 14.31 \\
LProto \cite{gupta2023class}                 & 70.03 & 73.14 & 30.25 & 13.71 & 5.43 & 74.47 & 15.34 \\
LMsc  \cite{audibert2024exploring}           & 69.09 & 71.54 & 30.14 & 13.41 & 5.27 & 74.15 & 14.82 \\
OML  \cite{imbriaco2022toward}                & 73.27 & 76.08 & 29.28 & 13.13 & 5.07 & 71.58 & \textbf{\textcolor{blue}{18.17}} \\
MulSupCon  \cite{zhang2024mulsupcon}         & 69.71 & 72.51 & 29.70 & 13.27 & 5.01 & 74.43 & 14.67 \\
Wg-MulSupCon \cite{zhang2024mulsupcon} & 70.32 & 73.50 & \textbf{\textcolor{blue}{30.34}} & 13.69 & 5.47 & \textbf{\textcolor{blue}{74.88}} & 15.17 \\
Sim-Dis Loss \cite{huang2024similarity}      & 67.74 & 70.01 & 30.12 & 13.55 & 5.31 & 72.67 & 13.37 \\
LReg  \cite{audibert2024survey}              & \textbf{\textcolor{blue}{75.23}} & \textbf{\textcolor{blue}{77.60}} & 29.64 & 13.60 & \textbf{\textcolor{blue}{5.96}} & 74.69 & \textbf{\textcolor{blue}{15.57}} \\
L-W/O-Reg  \cite{audibert2024survey}         & 70.88 & 73.68 & 30.25 & \textbf{\textcolor{blue}{13.72}} & 5.58 & 74.72 & 15.27 \\
\cline{1-8} 
\textbf{MARC (Ours) }                                  & 72.14 & 75.17 & 28.95 & \textbf{\textcolor{red}{14.28}} & \textbf{\textcolor{red}{6.85}} & 73.92 & 16.13 \\
\textbf{Wg-MARC (Ours)    }                      & \textbf{\textcolor{red}{77.71}} & \textbf{\textcolor{red}{80.20}} & 29.26 & 13.19 & 5.31 & \textbf{\textcolor{red}{77.88}} & \textbf{\textcolor{red}{19.13}} \\
\hline
\end{tabular}
\end{center}
\end{table*}


\begin{table*}[!t]
\begin{center} 
\caption{Retrieval results for ML-AID where \textcolor{red}{red} and \textcolor{blue}{blue} indicate the highest and second highest values respectively.}
\label{tab:results_mlaid}  
\fontsize{8pt}{12pt}\selectfont
\begin{tabular}{c|cc|ccc|cc}
\hline
\multirow{3}{*}{\textbf{Method}} & \multicolumn{2}{c|}{\multirow{1}{*}{\textbf{Cosine Similarity-Based}}} & \multicolumn{5}{c}{\multirow{1}{*}{\textbf{Relevance-Based Jaccard Similarity}}} \\ [2pt]
\cline{2-8}
 & \multirow{2}{*}{\textbf{mAP(sim)@5000}} & \multirow{2}{*}{\textbf{nDCG(sim)@100}} & \multicolumn{3}{c|}{\textbf{mAP Threshold}} & \multirow{2}{*}{\textbf{nDCG@100}} & \multirow{2}{*}{\textbf{wAP@100}} \\  [2pt]
\cline{4-6}
 &  &  & \textbf{Easy@0.4} & \textbf{Medium@0.6} & \multicolumn{1}{c|}{\textbf{Hard@0.8}} & &  \\  [2pt]
\hline 
SupCon ALL \cite{khosla2021supcon}            & 86.95 & 89.32 & 56.36 & 37.06 & 14.60 & 84.33 & 30.75 \\
SupCon ANY \cite{khosla2021supcon}            & 87.20 & 86.31 & 56.95 & 37.75 & \textbf{\textcolor{blue}{14.93}} & 82.22 & 26.90 \\
LBase   \cite{audibert2024exploring}          & 88.98 & 89.96 & 55.47 & 36.32 & 14.21 & 84.39 & 31.14 \\
LProto   \cite{gupta2023class}                & 88.41 & 89.64 & 56.18 & 36.94 & 14.73 & 83.98 & 30.28 \\
LMsc   \cite{audibert2024exploring}           & 89.10 & 89.77 & 56.33 & 36.96 & 14.51 & 83.83 & 29.71 \\
OML  \cite{imbriaco2022toward}                 & 93.13 & 93.52 & 57.90 & \textbf{\textcolor{blue}{38.20}} & \textbf{\textcolor{red}{15.00}} & 67.61 & 36.78 \\
MulSupCon \cite{zhang2024mulsupcon}           & 89.62 & 90.09 & 55.56 & 36.63 & 14.46 & 84.61 & 30.84 \\
Wg-MulSupCon \cite{zhang2024mulsupcon}  & 90.11 & 90.21 & 56.24 & 36.81 & 14.44 & 84.43 & 30.60 \\
Sim-Dis Loss \cite{huang2024similarity}       & 88.36 & 89.04 & 56.91 & 37.37 & 14.76 & 58.04 & 29.45 \\
LReg   \cite{audibert2024survey}              & 91.88 & 92.56 & 56.35 & 36.86 & 14.73 & 85.94 & 32.95 \\
L-W/O-Reg   \cite{audibert2024survey}         & 91.81 & 92.76 & 56.24 & 36.56 & 14.15 & \textbf{\textcolor{blue}{87.85}} & 36.98 \\
\cline{1-8} 
\textbf{MARC (Ours)  }                                  & \textbf{\textcolor{red}{95.77}} & \textbf{\textcolor{red}{95.90}} & \textbf{\textcolor{red}{59.89}} & \textbf{\textcolor{red}{38.86}} & 14.91 & 77.12 & \textbf{\textcolor{red}{43.84}} \\
\textbf{Wg-MARC (Ours)}                           & \textbf{\textcolor{blue}{94.46}} & \textbf{\textcolor{blue}{94.43}} & \textbf{\textcolor{blue}{58.20}} & 37.94 & 14.54 & \textbf{\textcolor{red}{89.84}} & \textbf{\textcolor{blue}{39.42}} \\
\hline
\end{tabular}
\end{center}
\end{table*}


\begin{table*}[!h]
\begin{center} 
\caption{Retrieval results for WHDLD dataset.
\textcolor{red}{Red} values indicate the highest performance among all methods, 
while \textcolor{blue}{blue} values indicate the second highest.}
\label{tab:results_whdld}  
\fontsize{8pt}{12pt}\selectfont
\begin{tabular}{c|cc|ccc|cc}
\hline
\multirow{3}{*}{\textbf{Method}} & \multicolumn{2}{c|}{\multirow{1}{*}{\textbf{Cosine Similarity-Based}}} & \multicolumn{5}{c}{\multirow{1}{*}{\textbf{Relevance-Based Jaccard Similarity}}} \\
\cline{2-8}
 & \multirow{2}{*}{\textbf{mAP(sim)@5000}} & \multirow{2}{*}{\textbf{nDCG(sim)@100}} & \multicolumn{3}{c|}{\textbf{mAP Threshold}} & \multirow{2}{*}{\textbf{nDCG@100}} & \multirow{2}{*}{\textbf{wAP@100}} \\
 
\cline{4-6}
 &  &  & \textbf{Easy@0.4} & \textbf{Medium@0.6} & \multicolumn{1}{c|}{\textbf{Hard@0.8}} & &  \\
\hline 
SupCon ALL \cite{khosla2021supcon}            & 90.87 & 92.77 & 82.86 & 66.66 & 42.26 & 78.06 & 61.54 \\
SupCon ANY  \cite{khosla2021supcon}           & 92.99 & 92.22 & 82.22 & 66.02 & 41.53 & 74.48 & 57.12 \\
LBase    \cite{audibert2024exploring}         & 93.23 & 92.46 & 82.81 & \textbf{\textcolor{blue}{67.32}} & \textbf{\textcolor{blue}{43.22}} & 76.59 & 59.78 \\
LProto  \cite{gupta2023class}                 & 90.62 & 92.47 & 82.91 & 66.45 & 41.75 & 73.90 & 56.50 \\
LMsc    \cite{audibert2024exploring}          & 93.87 & 92.59 & 82.21 & 66.34 & 41.98 & 77.20 & 60.48 \\
OML  \cite{imbriaco2022toward}                 & 92.65 & 91.22 & 81.43 & 64.78 & 40.36 & 70.80 & 52.37 \\
MulSupCon \cite{zhang2024mulsupcon}           & 93.46 & 92.23 & \textbf{\textcolor{blue}{83.15}} & 67.09 & 42.60 & 74.53 & 57.72 \\
Wg-MulSupCon \cite{zhang2024mulsupcon}  & 93.83 & 92.69 & 82.63 & 66.98 & 42.81 & 76.72 & 60.08 \\
Sim-Dis Loss \cite{huang2024similarity}       & 89.77 & 90.42 & 82.64 & 66.52 & 42.05 & 69.71 & 52.35 \\
LReg    \cite{audibert2024survey}             & 90.71 & 89.68 & 82.10 & 65.88 & 41.49 & 65.94 & 47.50 \\
L-W/O-Reg   \cite{audibert2024survey}         & 93.88 & 92.62 & 82.74 & 67.22 & \textbf{\textcolor{red}{43.26}} & 76.28 & 59.63 \\
\cline{1-8} 
\textbf{MARC (Ours) }                                   & \textbf{\textcolor{red}{94.41}} & \textbf{\textcolor{red}{93.72}} & 82.54 & 66.22 & 42.31 & \textbf{\textcolor{red}{79.42}} & \textbf{\textcolor{red}{63.24}} \\
\textbf{Wg-MARC (Ours)}                           & \textbf{\textcolor{blue}{94.14}} & \textbf{\textcolor{blue}{93.55}} & \textbf{\textcolor{red}{83.23}} & \textbf{\textcolor{red}{67.45}} & 43.10 & \textbf{\textcolor{blue}{78.20}} & \textbf{\textcolor{blue}{61.90}} \\
\hline\end{tabular}
\end{center}
\end{table*}

The retrieval results for the DLRSD, ML-AID, and WHDLD datasets are presented in 
Tables~\ref{tab:results_dlrsd}, \ref{tab:results_mlaid}, and 
\ref{tab:results_whdld}, respectively. Through these evaluations, comparisons 
across different dataset scales, label diversity levels, and scene complexity are 
enabled. Multiple baseline and state-of-the-art retrieval methods were evaluated, 
and retrieval behaviour based on both cosine similarity and jaccard relevance was 
assessed. These variations across datasets facilitate a comprehensive assessment 
of retrieval robustness and generalisation.

Across all three datasets, the proposed MARC is consistently shown to outperform 
competing methods in the majority of evaluation metrics, demonstrating strong 
generalisation ability. The results further indicate that more discriminative and 
semantically consistent embeddings are learned by MARC. Improvements for rare 
categories may be attributed to the adaptive weighting and dynamic-temperature 
mechanisms, through which balanced retrieval performance is maintained across both 
frequent and infrequent land-cover classes.

\subsubsection{DLRSD Results}
As shown in Table~\ref{tab:results_dlrsd}, both MARC and its weighted variant achieve strong and consistent performance across nearly all evaluation metrics. Weighted MARC attains clear gains in cosine similarity--based measures, while MARC performs best in relevance-based ranking, particularly at higher difficulty levels. The results indicate that the proposed approach maintains robust retrieval quality across varying similarity thresholds. Minor differences at the Easy and Medium levels suggest that MARC remains competitive even when other methods briefly outperform it, reflecting stable and balanced behavior across evaluation settings.
\subsubsection{ML-AID Results}
According to Table~\ref{tab:results_mlaid}, MARC attains the most consistent overall performance on the ML-AID dataset, confirming its strong generalization on large-scale and diverse remote sensing data. It leads across most similarity- and relevance-based measures, showing effective feature learning and ranking precision. Although a few baselines achieve slightly higher values in isolated metrics, MARC maintains a more uniform advantage across thresholds, highlighting its ability to capture complex semantic relations within multi-label aerial imagery.
\subsubsection{WHDLD Results}
From Table~\ref{tab:results_whdld}, MARC and Weighted MARC obtain the highest or near-highest results across all evaluation criteria. Weighted MARC achieves stronger retrieval performance at lower similarity thresholds, while MARC yields superior ranking quality. This complementary trend suggests that the weighted variant enhances precision in easier retrieval conditions, whereas MARC offers more reliable ranking at challenging levels. Overall, both versions of MARC sustain leading performance across different retrieval perspectives and dataset conditions.

\subsection{Qualitative Results}

\begin{figure*}[!ht]
\begin{center}
\captionsetup[subfloat]{justification=centering, singlelinecheck=false}

\begin{minipage}[c]{0.20\linewidth}
    \begin{center}
    \textbf{Query} \\
    \includegraphics[width=\linewidth]{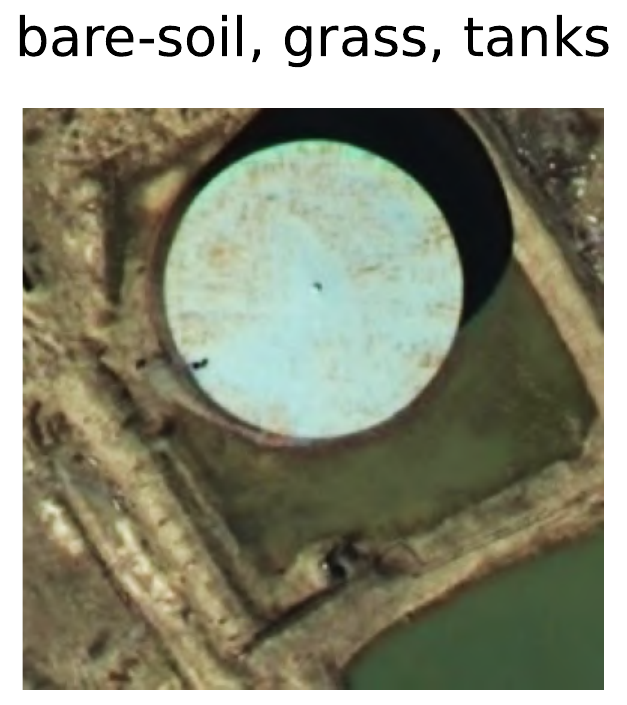}
    \end{center}
\end{minipage}
\begin{minipage}[c]{0.78\linewidth}

    \subfloat[MulSupCon]{
        \includegraphics[width=\linewidth]{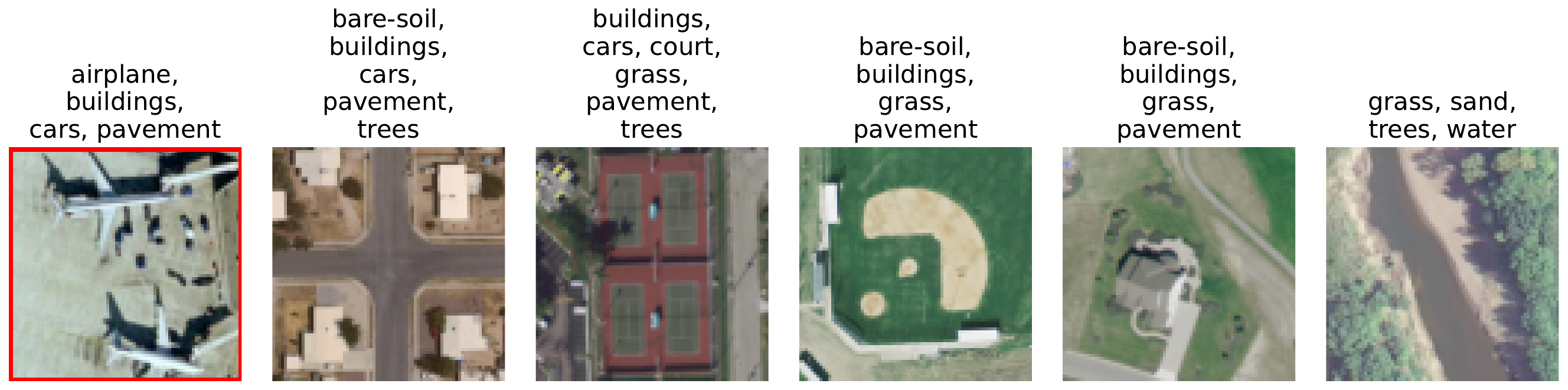}
    }

    \subfloat[LReg]{
        \includegraphics[width=\linewidth]{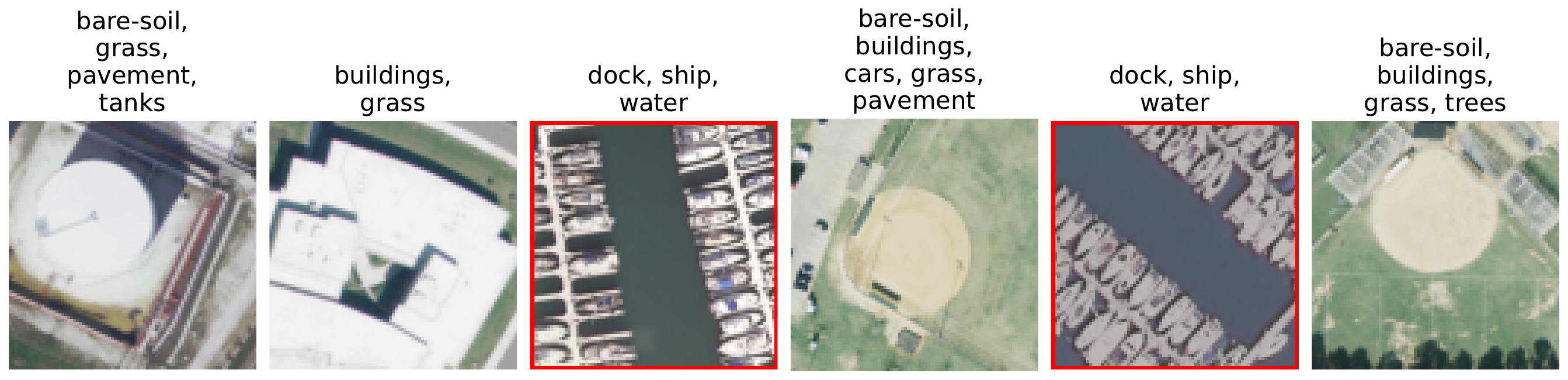}
    }

    \subfloat[MARC]{
        \includegraphics[width=\linewidth]{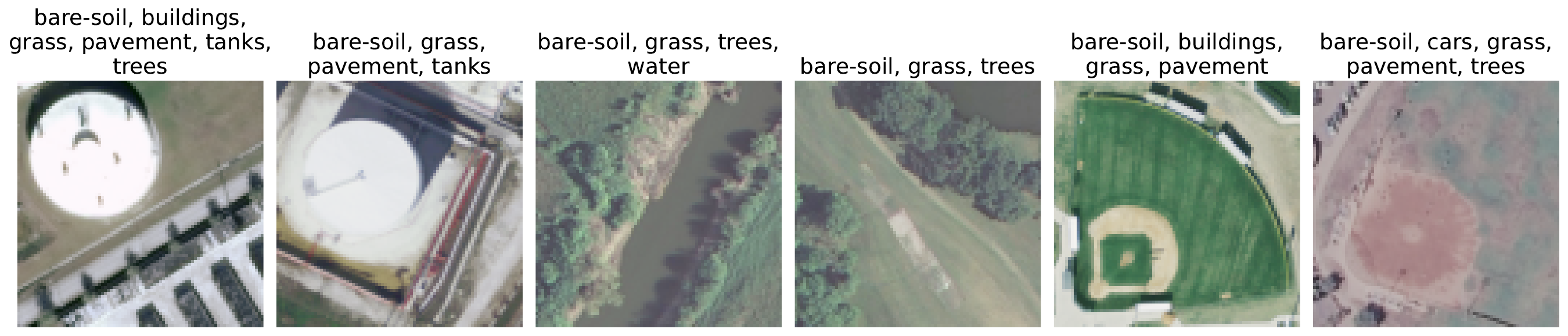}
    }

    \subfloat[Wg-MARC]{
        \includegraphics[width=\linewidth]{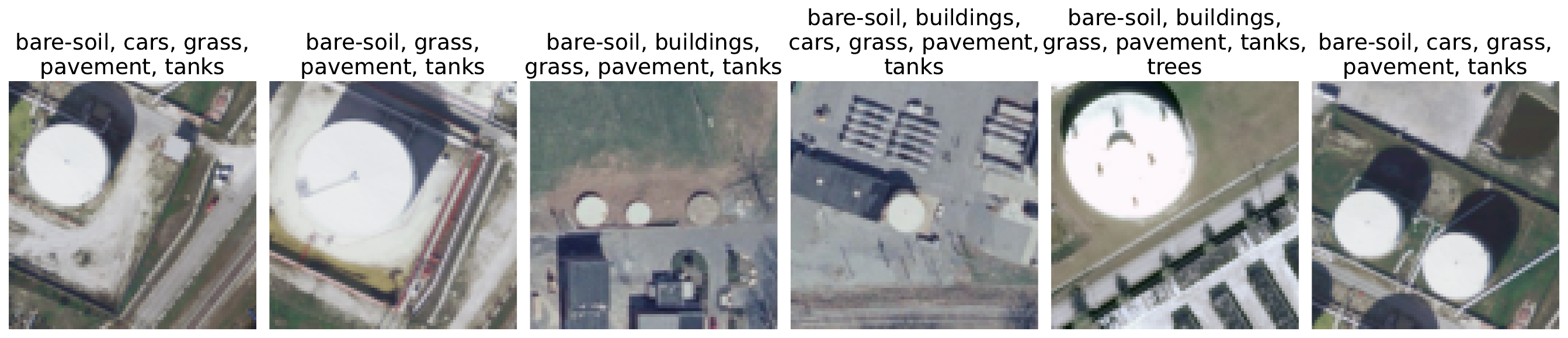}
    }

\end{minipage}

\caption{Top-6 retrieved images for the given query on the DLRSD dataset, comparing the best-performing methods (MulSupCon and LReg).}
\label{fig:dlrsd_retrieval}

\end{center}
\end{figure*}

A qualitative comparison of retrieval performance on the DLRSD dataset is 
presented to complement the quantitative evaluation. For a fixed query image, the 
top-6 retrieved samples produced by each method are shown in 
Figure~\ref{fig:dlrsd_retrieval}. Through this visualization, the effect of the 
loss function on the semantic relevance of the retrieved images can be observed. 
Methods that more effectively encode multi-label relationships tend to retrieve 
samples whose label sets closely correspond to those of the query image, whereas 
less effective methods often return visually similar yet semantically unrelated 
samples. These qualitative observations provide additional insight into the 
behaviour of the evaluated loss functions and help clarify performance differences 
that may not be fully captured by numerical metrics alone.

\subsection{Discussion}

The experimental results across all datasets demonstrate that MARC and its weighted variant consistently deliver improvements in multi-label remote sensing image retrieval. These gains are evident across cosine similarity–based metrics, relevance-based measures, and the qualitative retrieval examples. Several key observations can be drawn from these findings.

First, the consistent advantage of MARC across multiple evaluation metrics indicates that its adaptive mechanisms effectively address the limitations identified in the comparative study. In particular, the pairwise label reweighting component appears to mitigate the effects of class imbalance, enabling rare classes to exert greater influence during optimisation. This behaviour is reflected in the improved retrieval quality observed for infrequent land-cover categories, where baseline methods typically struggle.

Second, the dynamic temperature scaling mechanism is shown to improve the modelling of partial semantic similarity. By assigning temperature values based on label-overlap strength, the contrastive objective becomes more sensitive to nuanced relationships between samples. This results in embeddings that position semantically related images closer together while preventing visually similar but semantically unrelated samples from being mistakenly grouped. The observed improvements in nDCG and high-threshold mAP metrics particularly highlight the benefits of this adaptive formulation.

Third, the qualitative retrieval examples reinforce the quantitative trends by demonstrating clearer semantic alignment between query images and the top-ranked results produced by MARC-based models. These retrieved samples exhibit stronger agreement in label composition and scene characteristics compared to the outputs of competing loss functions. Such behaviour suggests that MARC produces a feature space that more faithfully represents the multi-label structure of remote sensing imagery.

Finally, the complementary behaviour observed between MARC and its weighted variant suggests that the proposed framework is flexible and can be tailored to different retrieval objectives. The weighted version performs especially well under lower similarity thresholds, whereas MARC itself yields superior ranking quality under stricter relevance conditions. This adaptability highlights the generalisability of the framework across diverse datasets and retrieval requirements.

Overall, the collective results confirm that MARC provides a robust and general-purpose solution for multi-label remote sensing image retrieval. The integration of frequency-aware weighting and adaptive temperature scaling leads to more discriminative, semantically coherent, and retrieval-effective feature representations compared to existing contrastive learning approaches.

\begin{figure*}[ht]
    \begin{center}
        \includegraphics[width=\linewidth]{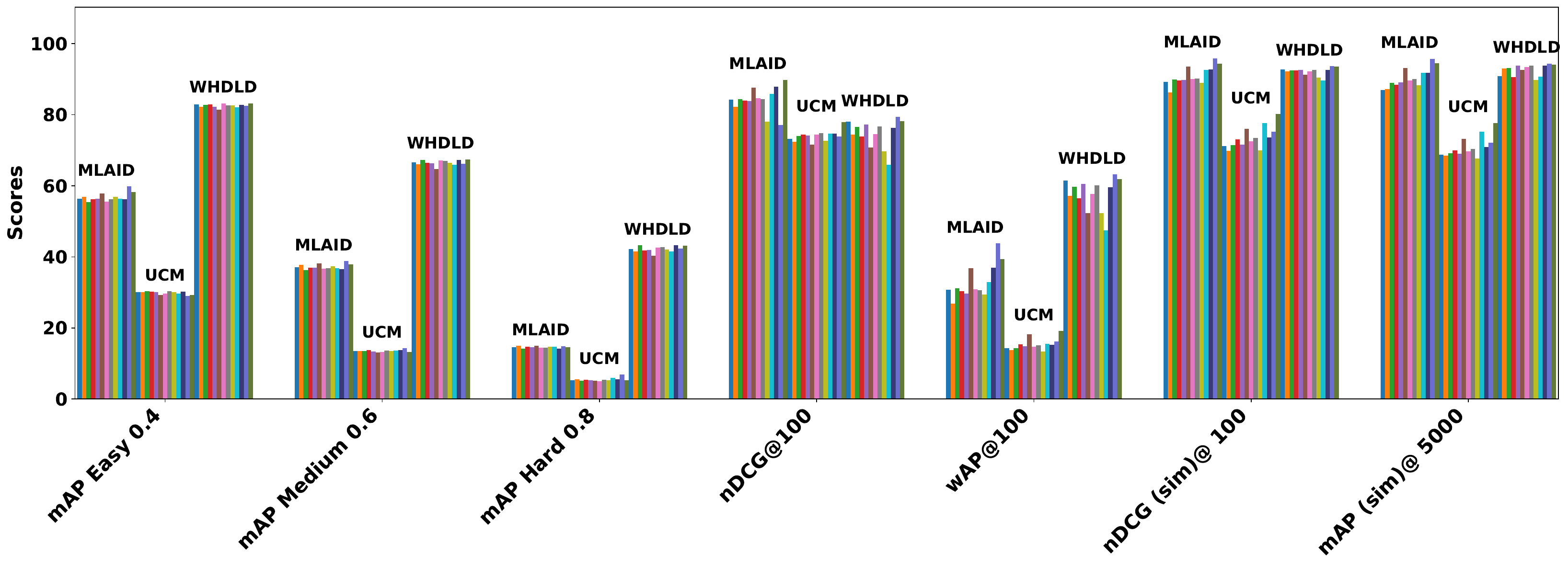}
    \end{center}
    \caption{Loss Function Comparison Across Metrics (All Datasets). 
    The evaluated loss functions are color-coded as follows: 
    SupCon ALL (\textcolor[HTML]{1f77b4}{\rule{6pt}{6pt}}), 
    SupCon ANY (\textcolor[HTML]{ff7f0e}{\rule{6pt}{6pt}}), 
    LBase (\textcolor[HTML]{2ca02c}{\rule{6pt}{6pt}}), 
    LProto (\textcolor[HTML]{d62728}{\rule{6pt}{6pt}}), 
    LMsc (\textcolor[HTML]{9467bd}{\rule{6pt}{6pt}}), 
    OML (\textcolor[HTML]{8c564b}{\rule{6pt}{6pt}}), 
    MulSupCon (\textcolor[HTML]{e377c2}{\rule{6pt}{6pt}}), 
    Wg-MulSupCon (\textcolor[HTML]{7f7f7f}{\rule{6pt}{6pt}}), 
    Sim-Dis Loss (\textcolor[HTML]{bcbd22}{\rule{6pt}{6pt}}), 
    LReg (\textcolor[HTML]{17becf}{\rule{6pt}{6pt}}), 
    L-W/O-Reg (\textcolor[HTML]{393b79}{\rule{6pt}{6pt}}), 
    MARC Loss (\textcolor[HTML]{5254a3}{\rule{6pt}{6pt}}),
    and Weighted MARC Loss (\textcolor[HTML]{637939}{\rule{6pt}{6pt}}).}
    \label{fig:all_datasets}
\end{figure*}

\subsection{Ablation Studies}

To better understand the contribution of different design choices in the proposed framework, 
Ablation experiments were performed by varying the loss parameters, 
specifically the weighting term $w_{ip}$ and the dynamic temperature $T_{ip}$, 
which is parameterized by the scaling factors $\alpha$ and $\beta$. 
Different global temperature values and learning rate schedulers have also been tested.

For these studies, the analysis was restricted to the \textbf{DLRSD dataset}. 
This choice was guided by two factors. 
First, the ablation study required extensive parameter sweeps and repeated training runs, 
making it computationally impractical to conduct across all datasets. 
Second, DLRSD is one of the most widely used benchmarks in the ML-RSIR literature 
and has been adopted in numerous prior studies. 
It therefore provides a representative and reliable basis for evaluating the design choices 
while keeping the experimental cost tractable.

The following subsections present detailed ablation results.

\subsubsection{\textbf{Effect of $\alpha$ and $\beta$ in $w_{ip}$ and $T_{ip}$}}

To investigate the interaction between the weighting and temperature scaling mechanisms, an ablation analysis was conducted to examine the influence of the scaling factors $\alpha$ and $\beta$ on the retrieval performance.

\begin{figure*}[!htbp] 
\begin{center}
\scalebox{0.85}{ 
\begin{tabular}{cc}   
\subfloat[$\alpha$--$\beta$ ($T_{ip}$ only)]{
    \includegraphics[width=0.5\textwidth]{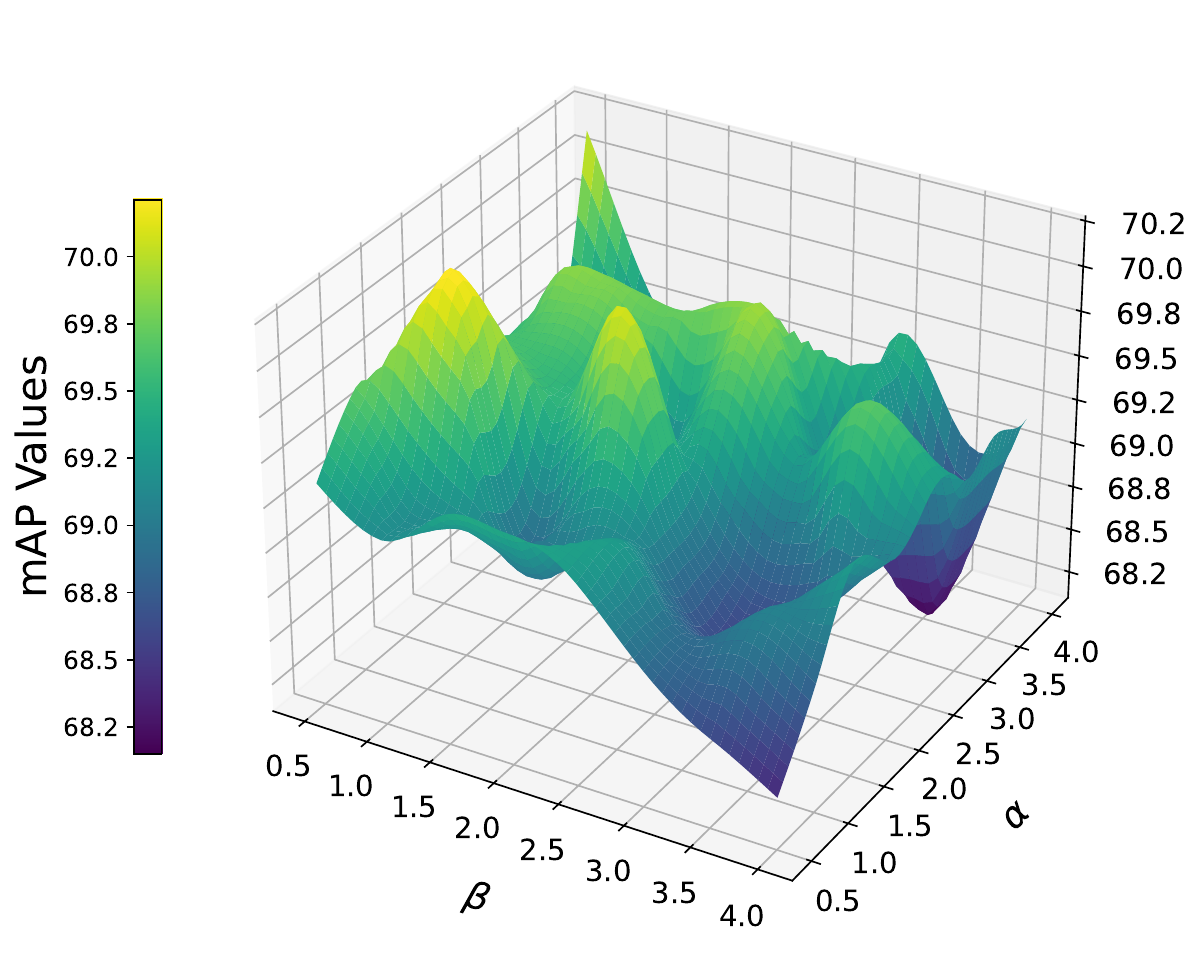}
}
\subfloat[$w_{ip}$ + $T_{ip}$ ($\alpha$ + $\beta$)]{
    \includegraphics[width=0.5\textwidth]{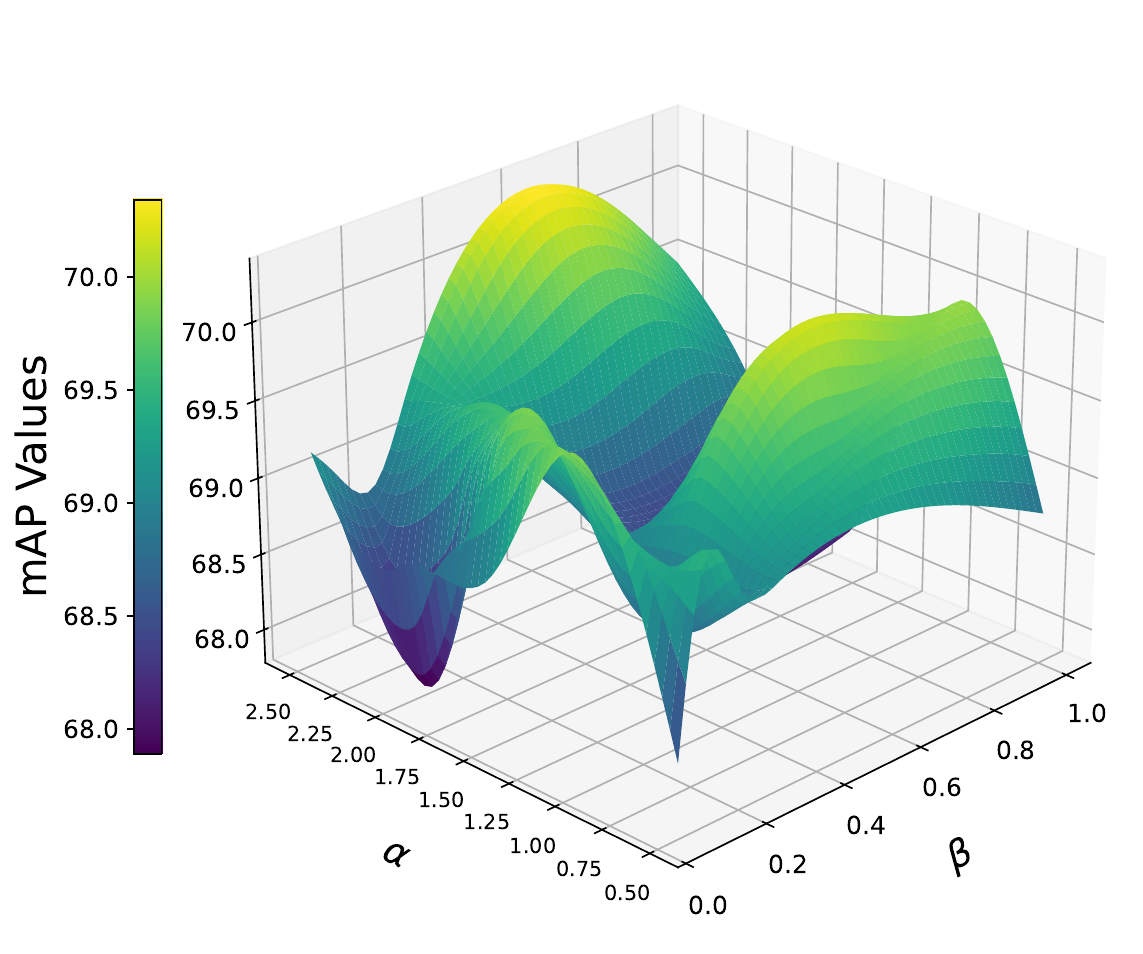}
}\\
\subfloat[$w_{ip}$ + $\alpha$]{
    \includegraphics[width=0.45\textwidth]{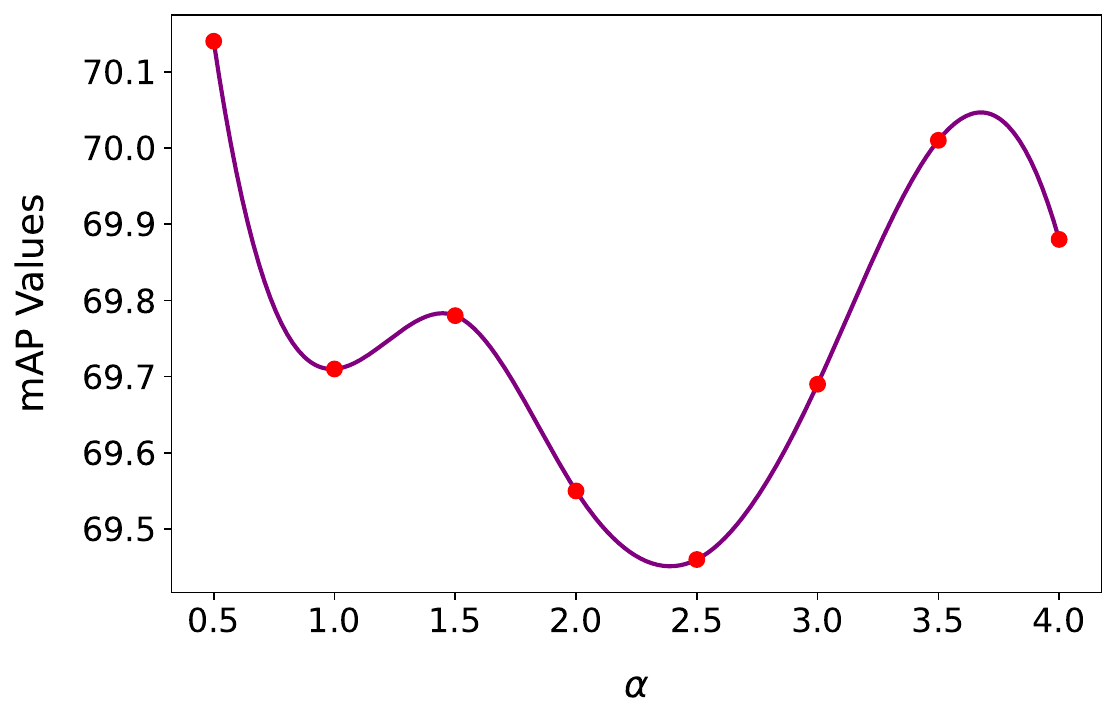}
}
\subfloat[$w_{ip}$ + $\beta$]{
    \includegraphics[width=0.45\textwidth]{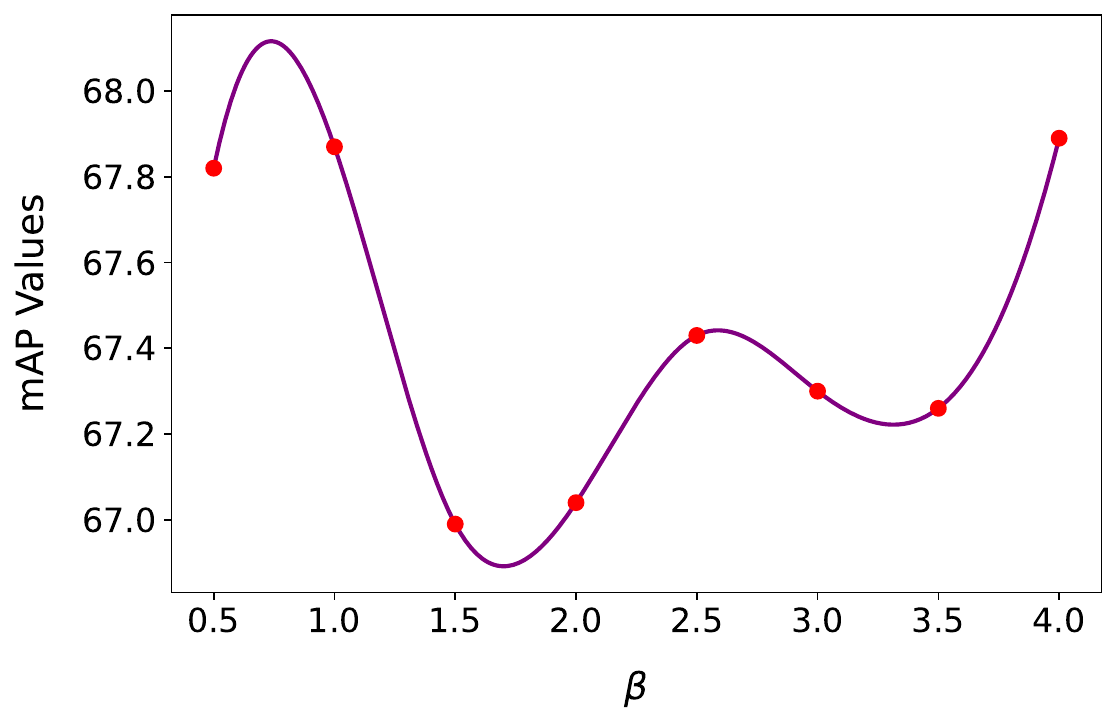}
}
\end{tabular}%
} 
\caption{Ablation study on the role of $T_{ip}$ (parameterized by $\alpha$ and $\beta$) 
and $w_{ip}$ in the proposed MARC loss.
Subfigures (a) and (b) show $\mathrm{mAP}$ performance as 3D surfaces 
(with colorbars indicating $\mathrm{mAP}$ in \%), 
while (c) and (d) show 2D $\mathrm{mAP}$ curves for varying $\alpha$ and $\beta$, respectively: 
(a) mAP Performance vs $\alpha$ and $\beta$ ($T_{ip}$ only), 
(b) mAP Performance vs $w_{ip}$ and $T_{ip}$ ($\alpha$ + $\beta$), 
(c) mAP vs $w_{ip}$ and $\alpha$, 
and (d) mAP vs $w_{ip}$ and $\beta$.}

\label{fig:ablation}
\end{center}
\end{figure*}

Four configurations were considered: 
(i) $\alpha$--$\beta$ in $T_{ip}$ without $w_{ip}$, 
(ii) the combined $w_{ip}$ + $T_{ip}$ case, 
(iii) $w_{ip}$ with $\alpha$, 
and (iv) $w_{ip}$ with $\beta$. 
The results are summarized in Fig.~\ref{fig:ablation}.

When either the weighting or temperature modules are removed, a noticeable reduction in retrieval performance is observed, confirming that semantic consistency and category balance are enhanced through these mechanisms. The design rationale of MARC is validated by these findings, and its ability to produce well-balanced representations is supported by the observed results.

\subsection*{Effect of $T_{ip}$ Only ($\alpha + \beta$)}  
As shown in Fig.~\ref{fig:ablation}(a), the variation of $\alpha$ and $\beta$ under the $T_{ip}$ only condition resulted in mAP values ranging from $68.1\%$ to $70.2\%$, with the highest value observed at $(\alpha{=}1.5,\beta{=}1.0)$. Moderate scaling was found to enhance representation learning, whereas extreme values reduced accuracy due to excessive temperature adjustment. These findings indicate that $T_{ip}$ provides a positive contribution when appropriately balanced.  
\subsection*{Effect of $w_{ip}$ and $T_{ip}$}  
In Fig.~\ref{fig:ablation}(b), the combination of $w_{ip}$ and $T_{ip}$ is shown to produce smoother and higher performance, with the maximum mAP recorded near $69.9\%$. The complementarity of the two modules is demonstrated, as $w_{ip}$ improves sample weighting and $T_{ip}$ adaptively regulates contrast strength, resulting in enhanced stability and retrieval accuracy.

\subsection*{Effect of $w_{ip}$ with $\alpha$}  
From Fig.~\ref{fig:ablation}(c), the mAP reached its highest value of $70.1\%$ when $\alpha$ was approximately $0.5$, after which a gradual decline was observed. Moderate values of $\alpha$ were found to provide the best balance between informative and redundant pairs, while larger $\alpha$ values placed excessive emphasis on strong pair relations, thereby reducing generalization.  

\subsection*{Effect of $w_{ip}$ with $\beta$}  
As shown in Fig.~\ref{fig:ablation}(d), $\beta$ exhibited a sharper sensitivity, with an optimal mAP of $67.9\%$ observed near $\beta{\approx}1.0$. Larger $\beta$ values were found to destabilize weighting, confirming that moderate scaling results in more consistent representations within the MARC framework.

\subsubsection{\textbf{Effect of Temperature Parameter}}

The temperature parameter $\tau$ is a critical factor in supervised contrastive learning, 
as it regulates the sharpness of similarity distributions. 
To evaluate its impact,An ablation study has been conducted with $\tau \in \{0.1, 0.2, 0.3\}$ 
across three benchmark datasets: DLRSD, ML-AID, and WHDLD. 
Very low values (e.g., $\tau < 0.09$) led to embedding collapse, where learned representations 
became indistinguishable, and were therefore omitted from the analysis.

\begin{figure*}[!ht]
\begin{center}

\subfloat[DLRSD]{%
    \includegraphics[width=0.32\textwidth]{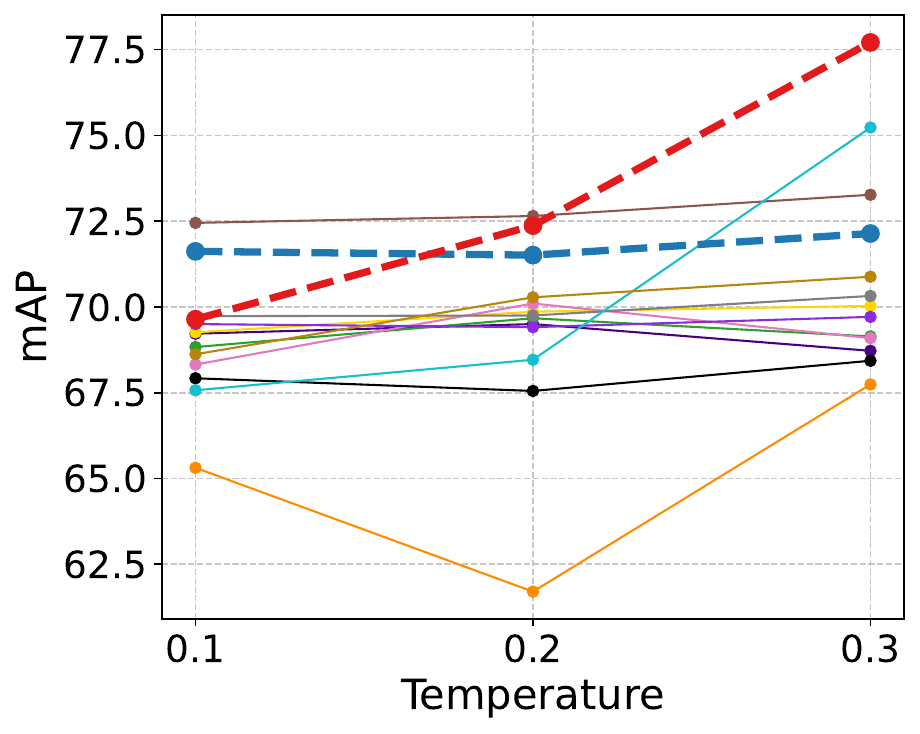}
}\hfill
\subfloat[ML-AID]{%
    \includegraphics[width=0.32\textwidth]{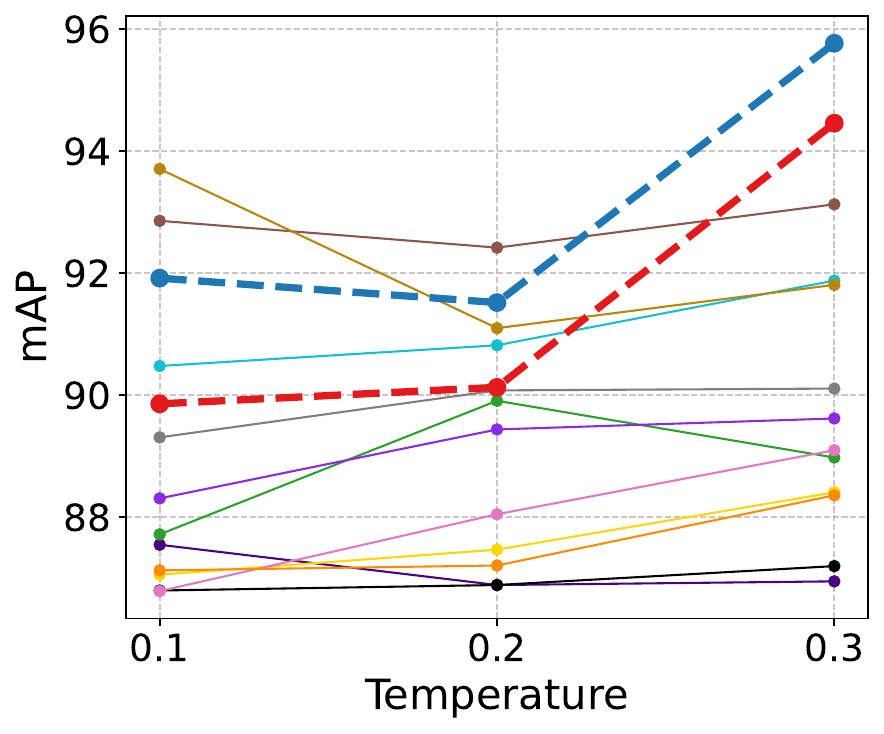}
}\hfill
\subfloat[WHDLD]{%
    \includegraphics[width=0.32\textwidth]{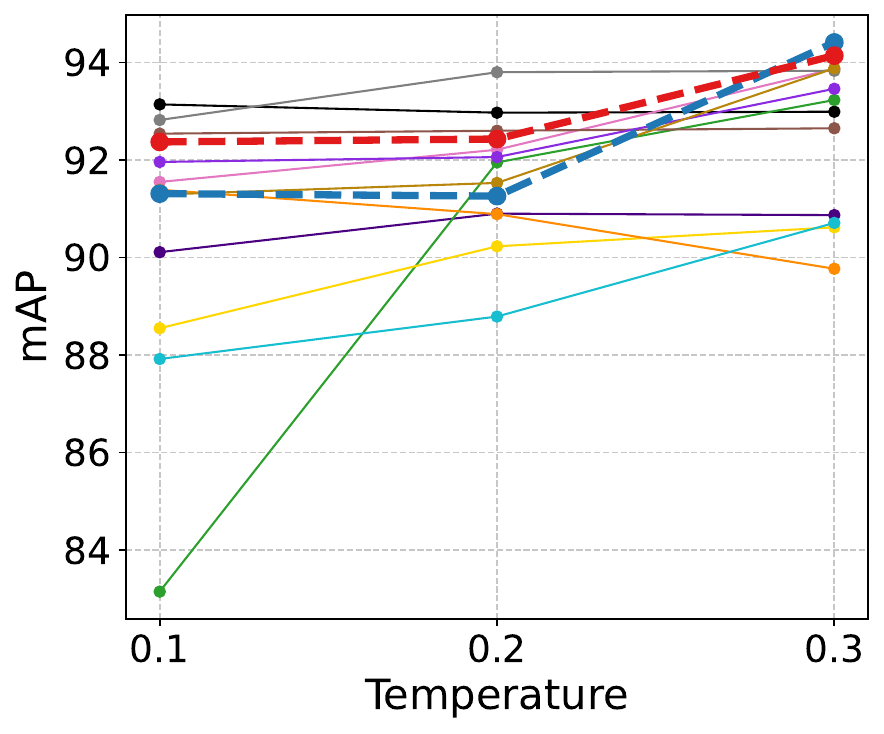}
}

\end{center}

\caption{Ablation study of the temperature parameter $\tau \in \{0.1,0.2,0.3\}$ vs. mAP across three datasets (DLRSD, ML-AID, and WHDLD). 
The evaluated loss functions are color-coded as follows: 
SupCon ALL (\textcolor[HTML]{4b0082}{\rule{6pt}{6pt}}), 
SupCon ANY (\textcolor[HTML]{000000}{\rule{6pt}{6pt}}), 
LBase (\textcolor[HTML]{2ca02c}{\rule{6pt}{6pt}}), 
LProto (\textcolor[HTML]{ffd700}{\rule{6pt}{6pt}}), 
LMsc (\textcolor[HTML]{e377c2}{\rule{6pt}{6pt}}), 
OML (\textcolor[HTML]{8c564b}{\rule{6pt}{6pt}}), 
MulSupCon (\textcolor[HTML]{8a2be2}{\rule{6pt}{6pt}}), 
Wg-MulSupCon (\textcolor[HTML]{7f7f7f}{\rule{6pt}{6pt}}), 
Sim-Dis Loss (\textcolor[HTML]{ff8c00}{\rule{6pt}{6pt}}), 
LReg (\textcolor[HTML]{17becf}{\rule{6pt}{6pt}}), 
L-W/O-Reg (\textcolor[HTML]{B8860B}{\rule{6pt}{6pt}}), 
MARC Loss (\textcolor[HTML]{1f77b4}{\rule{6pt}{6pt}}), 
and Weighted MARC Loss (\textcolor[HTML]{e31a1c}{\rule{6pt}{6pt}}).}

\label{fig:temperature_all}

\end{figure*}

As shown in Fig.~\ref{fig:temperature_all}, the mAP performance consistently improves as $\tau$ increases from $0.1$ to $0.3$.

This trend highlights the stabilizing effect of higher temperature on contrastive optimization, 
reducing over-clustering of positive pairs and enhancing retrieval quality.  
The highest performance across all datasets is achieved at $\tau=0.3$, which was consequently 
selected as the default for all subsequent experiments.  

This observation aligns with prior work in remote sensing contrastive learning, 
where $\tau=0.3$ was identified as a robust and stable benchmark that balances 
optimization dynamics and generalization performance\cite{liu2025dynamic,Amir2025Comparative}.

\subsubsection{\textbf{Effect of Pretrained Weights on ResNet-18 Backbone}}  

To analyze the role of pretrained weights, the proposed MARC and Weighted MARC losses were evaluated 
on ResNet-18 backbones with and without ImageNet initialization.  
Table\eqref{tab:resnet18_weights} summarizes the results across the DLRSD, ML-AID, and WHDLD datasets.  

Several patterns can be observed.  
First, pretrained weights consistently improve retrieval performance across all datasets and loss variants.  
For example, on ML-AID, MARC with weights achieves an mAP(sim)@5000 of $95.77\%$ compared to $91.84\%$ without weights.  
Second, the performance gap is more pronounced for Weighted MARC, indicating that pretrained initialization 
stabilizes the interaction between the loss components.  
Finally, although ResNet-18 is shallower than ResNet-50 or ResNet-101, the consistent improvements from 
pretrained initialization highlight its importance, particularly in remote sensing retrieval tasks 
where labeled training data is limited.

\begin{table*}[t]
\caption{Effect of pretrained ResNet-18 weights across three datasets for MARC and Weighted MARC. 
Results are reported with and without ImageNet weights.}
\label{tab:resnet18_weights}
\fontsize{7pt}{10pt}\selectfont
\begin{center}
\begin{tabular}{c|c|c|cc|ccc|cc}
\hline
\multirow{3}{*}{\textbf{Method}} & \multirow{3}{*}{\textbf{Loss}} &\multirow{3}{*}{\textbf{Pretrained}} & \multicolumn{2}{c|}{\multirow{1}{*}{\textbf{Cosine Similarity-Based}}} & \multicolumn{5}{c}{\multirow{1}{*}{\textbf{Relevance-Based Jaccard Similarity}}} \\
\cline{4-10}
 &  &  & \multirow{2}{*}{\textbf{mAP(sim)@5000}} & \multirow{2}{*}{\textbf{nDCG(sim)@100}} & \multicolumn{3}{c|}{\textbf{mAP Threshold}} & \multirow{2}{*}{\textbf{nDCG@100}} & \multirow{2}{*}{\textbf{wAP@100}} \\
\cline{6-8}
 &  &  &  &  & \textbf{Easy@0.4} & \textbf{Medium@0.6} & \multicolumn{1}{c|}{\textbf{Hard@0.8}} &  &  \\
\hline
\multirow{4}{*}{DLRSD} 
&  \multirow{2}{*}{MARC Loss}        & Yes & 72.14 & 75.17 & 28.95 & \textbf{\textcolor{red}{14.28}} & \textbf{\textcolor{red}{6.85}} & 73.92 & 16.13 \\ 
&                   & No  & 70.18 & 72.02 & \textbf{\textcolor{red}{29.89}} & 13.44 & 5.33 & 73.48 & 14.59 \\
\cline{2-10}
&  \multirow{2}{*}{Wg-MARC}   
& Yes & \textbf{\textcolor{red}{77.71}} & \textbf{\textcolor{red}{80.20}} & 29.26 & 13.19 & 5.31 & \textbf{\textcolor{red}{77.88}} & \textbf{\textcolor{red}{19.13}} \\
&                   & No  & 71.43 & 74.85 & 28.24 & 12.95 & 5.03 & 73.89 & 16.12 \\
\hline

\multirow{4}{*}{ML-AID} 
&  \multirow{2}{*}{MARC Loss}       & Yes & \textbf{\textcolor{red}{95.77}} & \textbf{\textcolor{red}{95.90}} & \textbf{\textcolor{red}{59.89}} & \textbf{\textcolor{red}{38.86}} & \textbf{\textcolor{red}{14.91}} & 77.12 & \textbf{\textcolor{red}{43.84}} \\
&                   & No  & 91.84 & 91.55 & 57.69 & 37.75 & 14.58 & 87.63 & 37.85 \\
\cline{2-10}
&  \multirow{2}{*}{Wg-MARC}      & Yes & 94.46 & 94.43 & 58.20 & 37.94 & 14.54 & \textbf{\textcolor{red}{89.84}} & 39.42 \\
&                   & No  & 89.26 & 89.83 & 54.85 & 35.86 & 14.25 & 83.87 & 31.09 \\
\hline

\multirow{4}{*}{WHDLD} 
&  \multirow{2}{*}{MARC Loss} & Yes & 94.41 & 93.72 & 82.54 & 66.22 & 42.31 & \textbf{\textcolor{red}{79.42}} & \textbf{\textcolor{red}{63.24}} \\
&                   & No  & \textbf{\textcolor{red}{94.50}} & \textbf{\textcolor{red}{93.89}} & 82.19 & 65.96 & 41.85 & 78.82 & 62.63 \\
\cline{2-10}
&  \multirow{2}{*}{Wg-MARC}        & Yes & 94.14 & 93.55 & \textbf{\textcolor{red}{83.23}} & \textbf{\textcolor{red}{67.45}} & \textbf{\textcolor{red}{43.10}} & 78.20 & 61.90 \\
&                   & No  & 92.69 & 92.48 & 83.22 & 67.02 & 42.58 & 75.76 & 58.85 \\
\hline

\end{tabular}
\end{center}
\end{table*}

\subsubsection{\textbf{Effect of Learning Rate}}  

\begin{figure*}[!htbp]
\begin{center}

\subfloat[DLRSD — MARC]{%
    \includegraphics[width=0.32\textwidth]{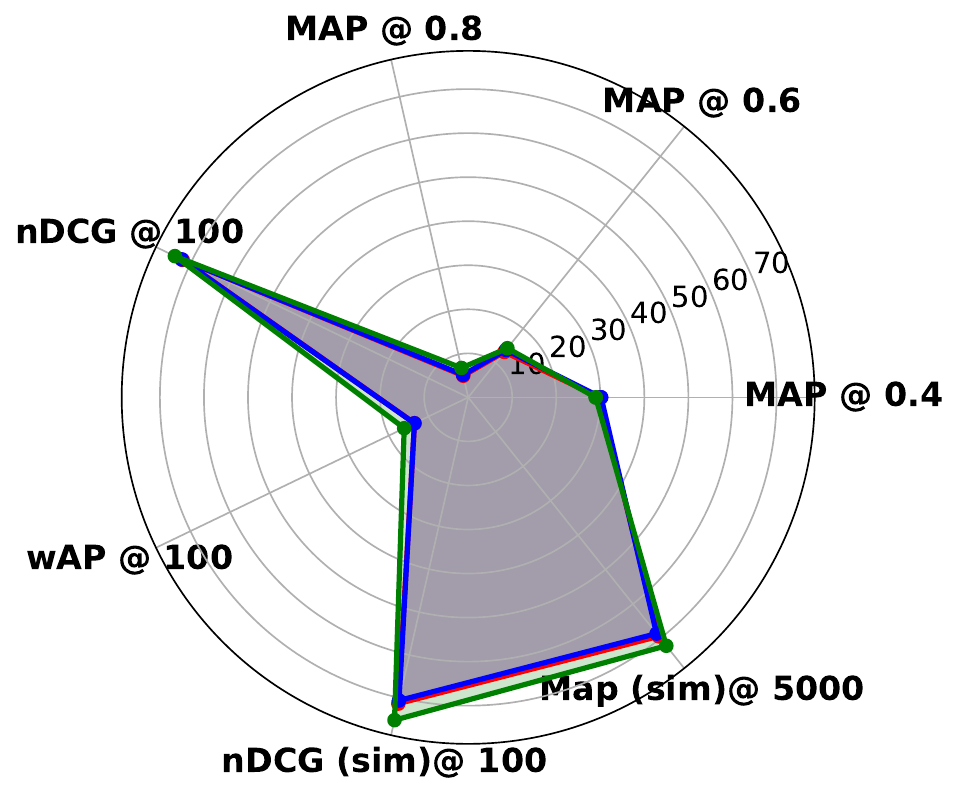}
}
\subfloat[WHDLD — MARC]{%
    \includegraphics[width=0.32\textwidth]{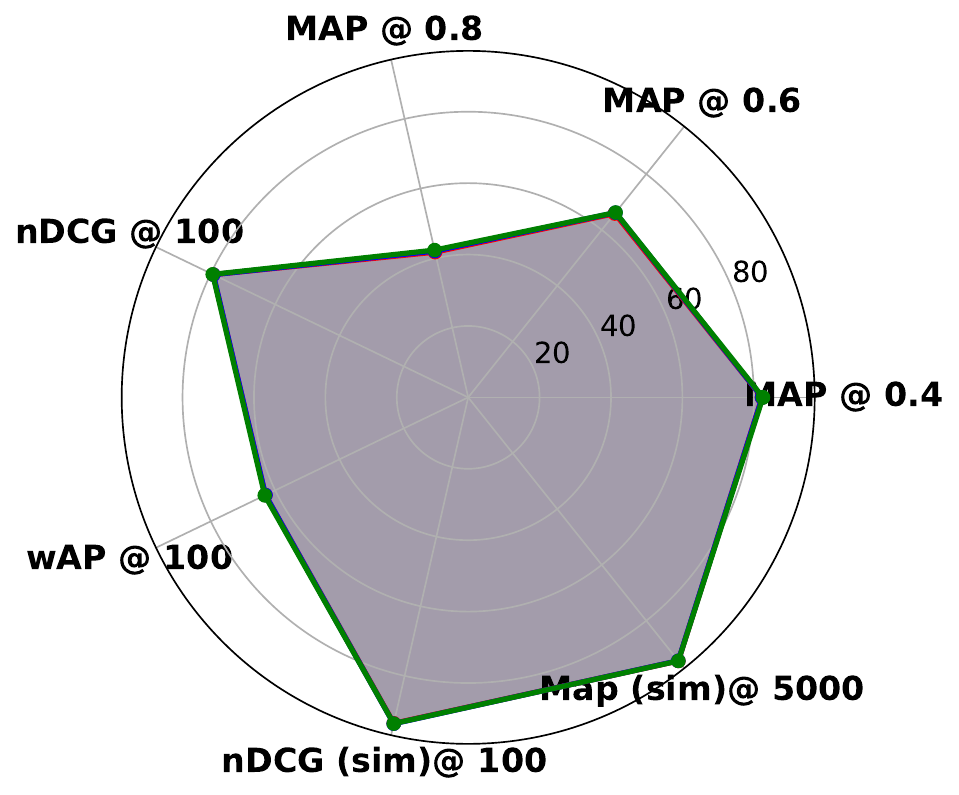}
}
\subfloat[ML-AID — MARC]{%
    \includegraphics[width=0.32\textwidth]{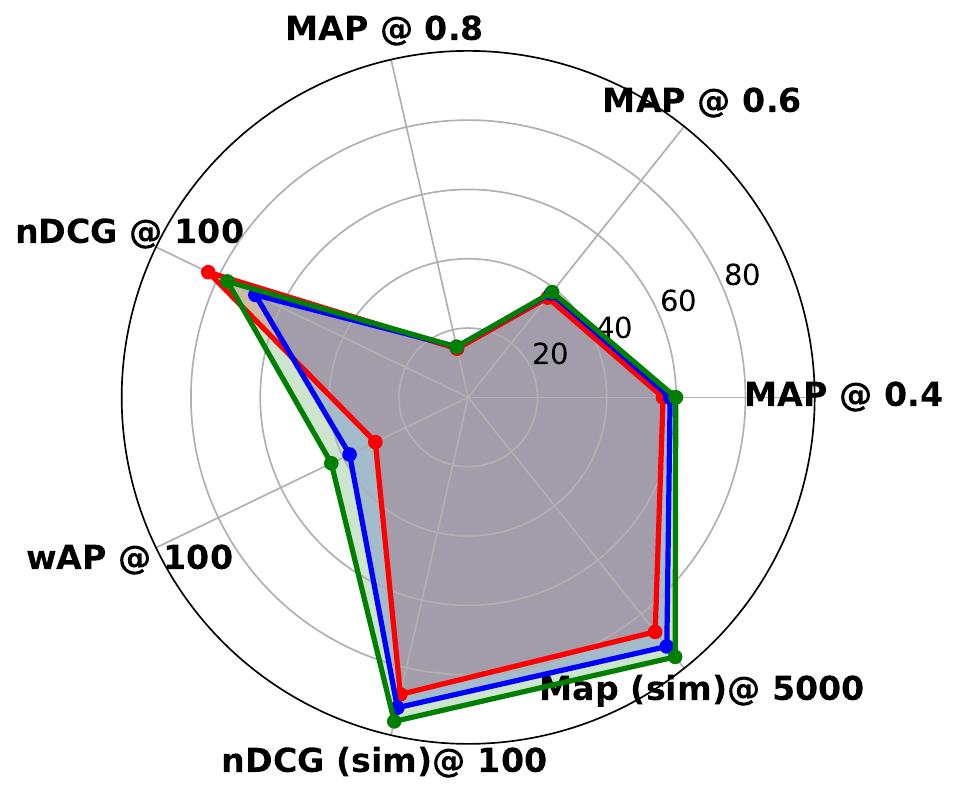}
}

\subfloat[DLRSD — Weighted MARC]{%
    \includegraphics[width=0.32\textwidth]{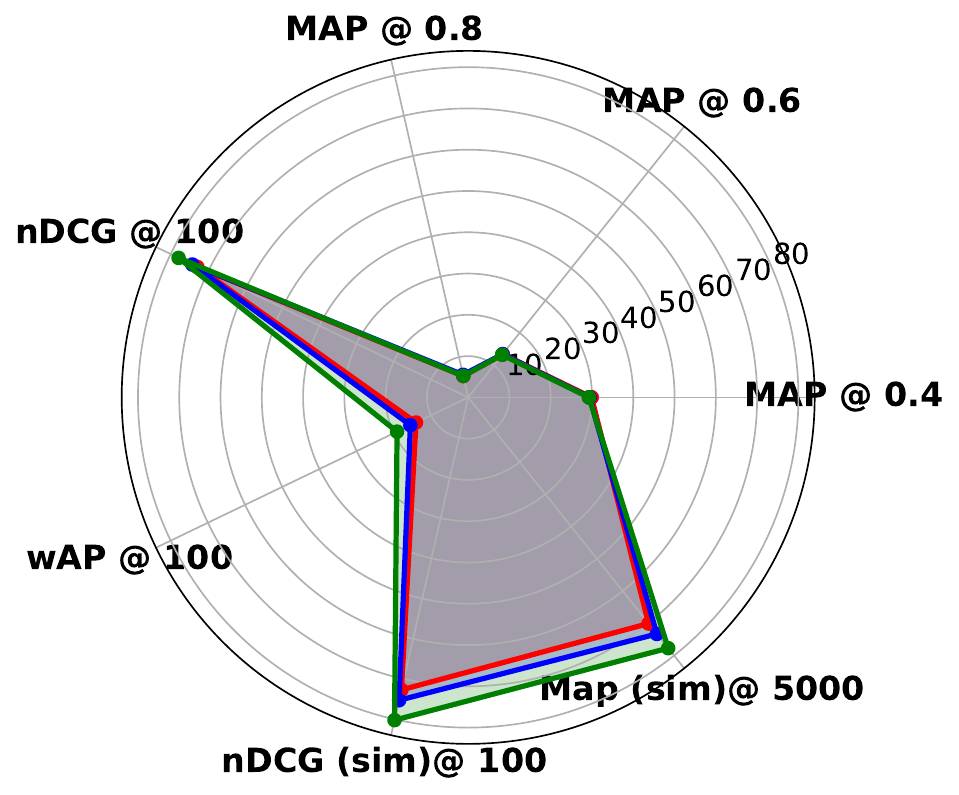}
}
\subfloat[WHDLD — Weighted MARC]{%
    \includegraphics[width=0.32\textwidth]{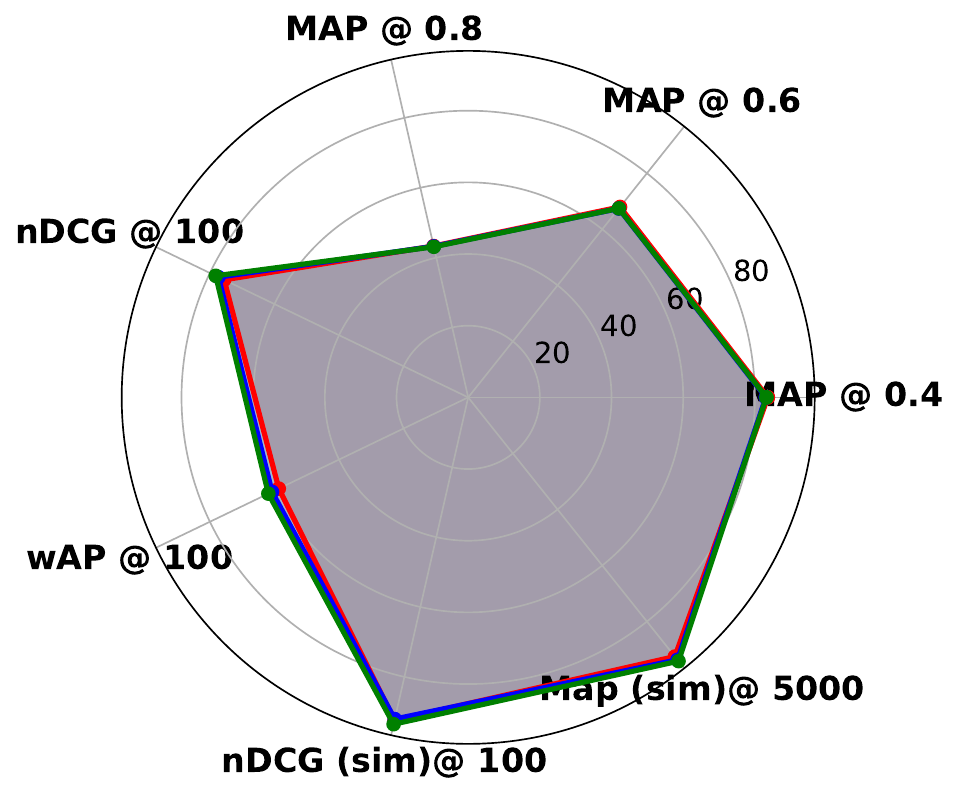}
}
\subfloat[ML-AID — Weighted MARC]{%
    \includegraphics[width=0.32\textwidth]{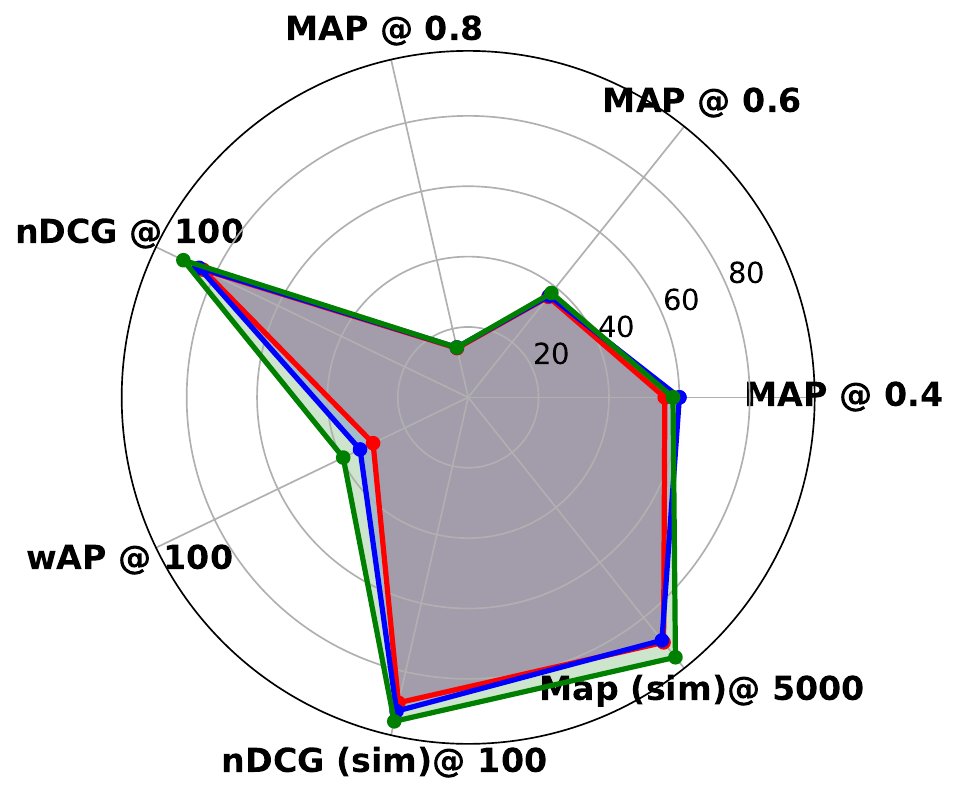}
}

\end{center}

\caption{Effect of learning rate on MARC and Weighted MARC across datasets. Smaller learning rates generally achieve the best retrieval accuracy.
Results are shown for
$\mathrm{LR}=0.1$ (\textcolor[HTML]{FF0000}{\rule{6pt}{6pt}}), 
$\mathrm{LR}=0.01$ (\textcolor[HTML]{0000FF}{\rule{6pt}{6pt}}), 
$\mathrm{LR}=0.001$ (\textcolor[HTML]{008000}{\rule{6pt}{6pt}}). 
}
\label{fig:lr_radar}

\end{figure*}

Learning rate plays a crucial role in determining convergence stability and final retrieval performance. 
Experiments were conducted with three values ($\text{LR} \in \{0.1, 0.01, 0.001\}$) across the DLRSD, ML-AID, and WHDLD datasets, 
considering both MARC and Weighted MARC losses. The results are shown in Fig.~\ref{fig:lr_radar}.

Smaller learning rates ($\text{LR}=0.001$) consistently yield the strongest retrieval results. 
For example, on ML-AID with MARC, $\text{mAP(sim)}@5000$ improves from $89.26\%$ at $\text{LR}=0.1$ to $95.77\%$ at $\text{LR}=0.001$. 
Similarly, Weighted MARC on DLRSD increases from $70.08\%$ at $\text{LR}=0.1$ to $77.71\%$ at $\text{LR}=0.001$.  

Intermediate learning rates ($\text{LR}=0.01$) often provide stable but slightly lower scores than $\text{LR}=0.001$. 
High learning rates ($\text{LR}=0.1$) generally degrade performance, particularly for Weighted MARC, 
as overly aggressive updates destabilize the balance between $w_{ip}$ and $T_{ip}$.  

The WHDLD dataset shows relative robustness, with all three learning rates achieving similar retrieval accuracy 
($\text{mAP(sim)}@5000 \approx 94\%$). This suggests that dataset scale and diversity mitigate sensitivity to the learning rate.  

Overall, the ablation confirms that $\text{LR}=0.001$ provides the most effective trade-off between convergence and retrieval accuracy, 
and this value is adopted in all subsequent experiments.

\section{Conclusion}
This study introduced MARC loss, 
a loss tailored for multi-label CBIR in remote sensing.  By integrating label-aware sampling, frequency-sensitive weighting, and dynamic temperature scaling, MARC effectively mitigates the challenges of imbalanced label distributions, semantic overlaps, and pairwise co-occurrence effects.
Comprehensive experiments on the DLRSD, ML-AID, and WHDLD datasets show that MARC consistently surpasses existing supervised contrastive variants across multiple metrics, 
producing discriminative, balanced, and semantically coherent embeddings.
The ablation analyses further confirm the role of each component, indicating that both the weighting and temperature mechanisms are essential for stable and high-quality retrieval performance and that MARC effectively achieves its design objectives.

Although MARC achieves strong performance, challenges remain in distinguishing visually similar classes (e.g., river vs. runway). Future work will focus on modeling higher-order label correlations, exploring multi-modal extensions, and adapting MARC for semi-supervised or weakly labeled settings.

\bibliographystyle{unsrt}  
\bibliography{references}
\end{document}